\documentclass{article}

\PassOptionsToPackage{numbers,square,comma,sort&compress}{natbib}

\usepackage[preprint]{neurips_2026}

\setcitestyle{numbers,square,comma,sort&compress}

\usepackage[utf8]{inputenc} 
\usepackage[T1]{fontenc}    
\usepackage{xurl}
\usepackage{booktabs}       
\usepackage{amsfonts}       
\usepackage{nicefrac}       
\usepackage{microtype}      

\usepackage{graphicx}
\usepackage{subcaption}
\usepackage{makecell}
\usepackage{multirow}
\usepackage{algorithm}
\usepackage{algorithmic}

\usepackage{amsmath}
\usepackage{amssymb}
\usepackage{mathtools}
\usepackage{amsthm}
\usepackage{enumitem}

\usepackage[font=small, labelfont=bf]{caption}

\usepackage{placeins}
\usepackage{float}

\usepackage{tabularx}
\usepackage{array}

\usepackage[table]{xcolor}

\definecolor{amsPurple}{HTML}{674EA7}       
\definecolor{amsLavender}{HTML}{F4F0FA}     
\definecolor{amsLavenderDark}{HTML}{E7DCF5} 

\definecolor{amsBestPurple}{HTML}{A824FF}   

\definecolor{sepGray}{HTML}{EEF2FF}         
\definecolor{sepGrayDark}{HTML}{E0E7FF}     

\definecolor{amsTextGray}{HTML}{475569}     
\definecolor{amsRed}{HTML}{E11D48}          
\definecolor{amsGray}{HTML}{818EA3}         
\colorlet{amsviolet}{amsLavender}

\newcommand{\drop}[1]{{\color{amsRed}\fontsize{7pt}{8pt}\selectfont $\downarrow$#1}}

\newcommand{\neut}[1]{{\color{amsGray}\fontsize{7pt}{8pt}\selectfont \textbf{-}}}

\newcommand{\amsgain}[1]{%
  {\textcolor{amsBestPurple}{\scriptsize\ensuremath{^{\uparrow #1}}}}%
}

\newcommand{\best}[1]{\textcolor{amsBestPurple}{\textbf{#1}}}

\newcolumntype{Y}{>{\centering\arraybackslash}X}
\newcolumntype{Z}{>{\raggedright\arraybackslash}X}
\newcolumntype{L}[1]{>{\raggedright\arraybackslash}p{#1}}

\usepackage{hyperref}
\hypersetup{
    colorlinks=true,
    linkcolor=black,
    citecolor=amsBestPurple,
}

\usepackage[capitalize,noabbrev]{cleveref}

\title{Adaptive Mass-Segmented KV Compression for Long-Form Reasoning}

\author{%
  \textbf{Junzhe Yang}\textsuperscript{1,2},
  \textbf{Xiaoyu Shen}\textsuperscript{2}\thanks{Corresponding author.}
  \\
  \\
  \textsuperscript{1} Shanghai Jiao Tong University \\
  \textsuperscript{2} Institute of Digital Twin, Eastern Institute of Technology, Ningbo \\
  \texttt{yang\_j@sjtu.edu.cn}, \texttt{xyshen@eitech.edu.cn}
}

\begin{document}

\maketitle

\begin{abstract}
The linear growth of the Key-Value (KV) cache is a critical bottleneck in long-form LLM inference. Existing KV compression methods mitigate this by evicting tokens based on importance scores. However, we show that their reliance on global Top-$k$ selection triggers \emph{Region Wipe-out}: the severe eviction of contiguous reasoning blocks that derails logical coherence. To address this, we propose Adaptive Mass-Segmented (AMS) KV Compression, a framework that shifts the paradigm from token-level competition to region-aware quota allocation. AMS adaptively partitions the KV cache based on the spatial distribution of attention mass, ensuring structurally vital reasoning segments receive guaranteed memory quotas. To ensure stability during iterative decoding, an EMA-based smoothing mechanism is incorporated to prevent jitter in segment boundaries. Crucially, AMS is a universal plug-and-play layer that is orthogonal to existing scorers. It can be seamlessly integrated into representative methods such as TOVA, Expected Attention, KeyDiff, R-KV and TriAttention. AMS is also system-compatible with modern paged-KV serving frameworks such as vLLM, supporting efficient gather-and-compact KV execution without introducing additional steady-state attention overhead. Extensive experiments across a diverse suite of tasks, including mathematical reasoning (\textsc{Math500}, \textsc{AIME}, \textsc{GSM8K}), code completion, open-domain QA, and sparse retrieval, demonstrate that AMS consistently mitigates structural fragmentation and boosts model performance. Code is available at: \url{https://github.com/EIT-NLP/AdaptiveMassSegment}

\end{abstract}

\section{Introduction}
\label{sec:intro}


While LLMs excel at complex tasks, the linear growth of the KV cache has become a major bottleneck in inference~\cite{h2o,kwon2023vllm,kvquant}. This issue is particularly pronounced in long-form reasoning, where generating extended Chain-of-Thought (CoT) sequences continuously expands the cache~\cite{wei2022chain,rkv,rpc,triattention_paper}. As decoding is fundamentally IO-bound, repeatedly loading these growing KV states from HBM creates a memory wall that significantly degrades latency \cite{dao2022flashattention,kwon2023vllm}.

To mitigate this cost, decoding-time KV compression offers a practical mechanism for bounded memory inference~\cite{streamingllm,h2o,oren-etal-2024-transformers}. These methods periodically compress the KV cache during generation by performing on-the-fly eviction of less important entries~\cite{pyramidkv,omnikv,lacache,adakv,duoattention,headkv,razorattention,snapkv,sablock,clusterkv,protokv,treekv,heterocache}. At each compression step, tokens are scored based on importance metrics such as accumulated or expected attention \cite{streamingllm,h2o,oren-etal-2024-transformers,expectedattn,chunkkv,adakv,restkv}, geometric variation \cite{keydiff,triattention_paper}, or reasoning-aware criteria \cite{rkv,rpc}. Despite the diversity in metrics, these approaches often rely on a \emph{budgeted token-level competitive} paradigm, where individual tokens across the entire context compete globally for a fixed number of slots via Top-$k$ selection~\cite{h2o,oren-etal-2024-transformers,expectedattn,keydiff,scope,restkv,gkv}.

However, we find that this token-level competition introduces a critical structural weakness under aggressive compression. Top-$k$ selection naturally favors highly salient ``heavy hitter'' tokens~\cite{h2o,liu2023scissorhands}, while distributed but contextually important reasoning spans receive insufficient protection. As a result, entire intermediate reasoning regions can be removed from the cache, a failure mode we term \emph{Region Wipe-out}. Related forms of KV fragmentation and context loss have also been observed in prior work~\cite{kvcompress,scope}. Region Wipe-out breaks the continuity of reasoning trajectories and introduces temporal gaps that impair logical consistency~\cite{liu2024lost}, leading to several characteristic behaviors such as \emph{(1) Problem Drifting}, where the model silently alters key constraints; \emph{(2) Premature Overturning}, where valid intermediate conclusions collapse after supporting context is removed; and \emph{(3) Repetition Collapse}, where severe context starvation drives the model into degenerate loops \cite{holtzman2020curious}.~\footnote{These phenomena are illustrated in Figure~\ref{fig:case-study} and further analyzed in Appendix~\ref{app:mechanistic-diagnostics}.} Several prior works attempt to reduce fragmentation through chunk- or block-based compression~\cite{chunkkv,kvcompress,treekv}. While such approaches partially alleviate direct token competition, their partition structures remain largely fixed and cannot adapt to the evolving distribution of reasoning importance. Consequently, they often fail to adequately protect dense reasoning spans. 

We propose Adaptive Mass-Segmented (AMS) KV compression, a framework that reformulates KV compression from \emph{token-level direct competition} into \emph{region-aware quota allocation}. Instead of using importance scores solely as individual survival signals, AMS interprets them as a global spatial distribution over the reasoning trajectory. Our framework first constructs adaptive segments based on the distribution of attention mass, then allocates retention quotas with minimum preservation guarantees for each segment before performing local token selection. This \emph{allocate-then-score} paradigm explicitly shields structurally important regions from being entirely wiped out.
Moreover, because decoding-time compression is repeatedly applied throughout generation, allocation decisions based purely on instantaneous attention signals can fluctuate across compression events~\cite{restkv,gkv,fragilitykv}. To reduce quota jitter across compression events, AMS introduces a lightweight history-aware regional mass estimation mechanism based on exponential moving average (EMA) smoothing~\cite{restkv,gkv}, applying temporal stabilization at the allocation level rather than modifying the base scorer.~\footnote{As shown in Table~\ref{tab:ablation_main}, removing EMA credit degrades accuracy, especially at larger cache budgets.}


Extensive experiments demonstrate the effectiveness of AMS across diverse reasoning and generation tasks, including mathematical reasoning (\textsc{Math500}, \textsc{AIME}, \textsc{GSM8K}), code completion, open-domain question answering, and sparse retrieval. Notably, AMS delivers its strongest gains under constrained KV budgets: on \textsc{Math500}, AMS-Expected improves over AdaKV-ExpE2 by up to $+16.0$ points, surpasses uncompressed Full KV at $T_{\mathrm{keep}}=512$ and $1024$, and further boosts the strong TriAttention scorer by up to $+6.4$ points. A key advantage of AMS lies in its modularity: rather than replacing existing token importance estimators, it serves as an orthogonal allocation layer that can be seamlessly integrated into existing KV compression pipelines such as TOVA~\cite{oren-etal-2024-transformers}, Expected Attention~\cite{expectedattn}, TriAttention~\cite{triattention_paper}, KeyDiff~\cite{keydiff}, and R-KV~\cite{rkv}, consistently improving their performance without architectural modifications (Table~\ref{tab:math500-universality}). Beyond accuracy, AMS also exhibits strong systems practicality. Unlike masking-based methods such as AdaKV-ExpE2 that retain the full KV cache in memory, AMS follows a gather-and-compact design that maintains the same physical cache footprint as standard gather-based compressors while achieving substantially stronger performance. AMS is naturally compatible with vLLM-style paged KV serving: the selector produces head-wise keep indices, while the runtime materializes them through paged-KV compaction and block-table replacement, allowing standard compact-cache attention without additional steady-state attention overhead (Appendix~\ref{app:vllm-compat}). Empirically, AMS matches or improves the latency of existing gather-based policies, while also reducing repetition collapse during free-form reasoning, leading to faster end-to-end generation under the same KV budget (Appendix~\ref{app:latency}).

Our contributions are as follows: (1) We identify \emph{Region Wipe-out} as a fundamental failure mode in KV cache compression, revealing how token-level Top-$k$ selection disrupts long-form reasoning under tight memory budgets. (2) We propose \emph{Adaptive Mass-Segmented (AMS) KV Compression}, a region-aware allocate-then-score framework that combines adaptive segmentation, segment-wise quota allocation, and EMA-based stabilization to preserve coherent reasoning trajectories during repeated compression. (3) We show that AMS serves as a general plug-and-play allocation layer that can be seamlessly integrated into diverse KV compression methods, consistently improving both effectiveness and robustness under constrained memory budgets. (4) We demonstrate that AMS is not only accurate but also system-efficient, supporting memory-efficient gather-based KV compaction, compatibility with vLLM-style paged serving, and latency-efficient long-context decoding without sacrificing practical deployment efficiency.

\section{Related Work}

\paragraph{Decoding-time KV eviction and training-free scoring.}
Decoding-time KV compression bounds memory by evicting cached tokens online, typically combining locality with globally important tokens such as sinks or heavy hitters \cite{streamingllm,h2o}. Many training-free methods score tokens by attention, reconstruction, geometry, or reasoning-aware proxies, including TOVA, KeyDiff, SnapKV, expected-attention scoring, TriAttention, R-KV, and RPC \cite{oren-etal-2024-transformers,keydiff,snapkv,expectedattn,triattention_paper,rkv,rpc}. However, retention is still often implemented as global or per-head top-$k$ selection, which lacks explicit protection for contiguous temporal regions and can be brittle under tight budgets and repeated recompression \cite{scope,restkv,gkv}.

\paragraph{Adaptive budgeting and structure-aware compression.}
Prior work allocates KV budgets non-uniformly across layers or heads, such as PyramidKV, OmniKV, LaCache, AdaKV, DuoAttention, HeadKV, and RazorAttention \cite{pyramidkv,omnikv,lacache,adakv,duoattention,headkv,razorattention}. Other methods organize caches into chunks, blocks, clusters, prototypes, or trees, including ChunkKV, SABlock, ClusterKV, ProtoKV, TreeKV, and HeteroCache \cite{chunkkv,sablock,clusterkv,protokv,treekv,heterocache}. These approaches improve budget allocation across layers, heads, or coarse cache units, but they do not explicitly allocate quotas to adaptive temporal regions within each head. In contrast, AMS introduces \emph{attention-derived adaptive segments} along the sequence and assigns explicit region-wise quotas before token-level scoring.

\paragraph{Stability and systems.}
Repeated recompression can destabilize context usage, motivating history-aware scoring such as ReST-KV and G-KV \cite{restkv,gkv}. AMS instead makes the \emph{allocation} history-aware, using EMA credit to smooth adaptive segments and quotas while keeping the scorer unchanged. Recent long-output reasoning systems such as LongFlow, Lethe, and ThinKV improve KV efficiency through fused eviction, layer/time-adaptive pruning, or thought-level quantization--eviction; AMS is orthogonal, providing scorer-agnostic within-head temporal quota allocation \cite{su2026longflow,zeng2025lethe,ramachandran2025thinkv}.~\footnote{We discuss paged-KV compatibility with vLLM~\cite{kwon2023vllm} in Appendix~\ref{app:vllm-compat}.}

\begin{figure*}[t]
  \centering
  \includegraphics[width=\textwidth]{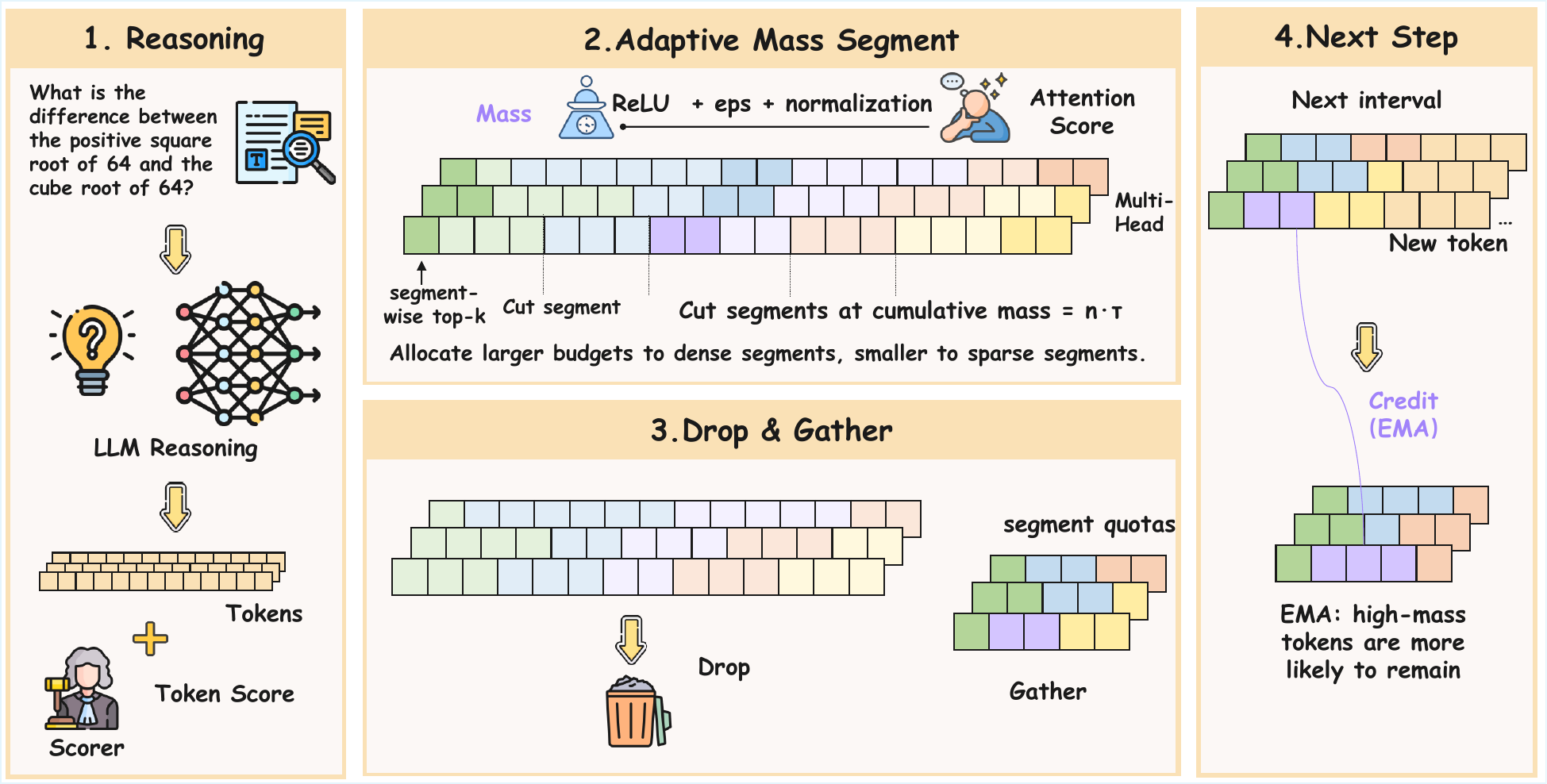}
  \caption{\textbf{Overview of AMS for decoding-time KV compression.}
  (1) Given a long chain-of-thought reasoning process, the base model generates
  tokens and a training-free scorer assigns token-level importance scores.
  (2) AMS converts recent attention-based usage into a normalized quality mass,
  partitions the sequence into adaptive segments based on cumulative mass, and
  allocates larger quotas to dense segments and smaller quotas to sparse ones.
  (3) Within each segment, AMS performs segment-wise top-$k$ selection under the
  segment quota, drops evicted tokens, and gathers the retained KV entries into
  a compact cache. (4) At the next compression event, AMS reuses an EMA-based
  credit signal that accumulates past mass, making consistently high-mass
  tokens more likely to remain and stabilizing selection across events.}
  \label{fig:method-overview}
\end{figure*}

\section{Method}
\label{sec:method}
\subsection{Fundamental Setting}
\begin{figure}[!t]
  \centering
  \begin{subfigure}[t]{0.49\columnwidth}
    \centering
    \includegraphics[width=\linewidth]{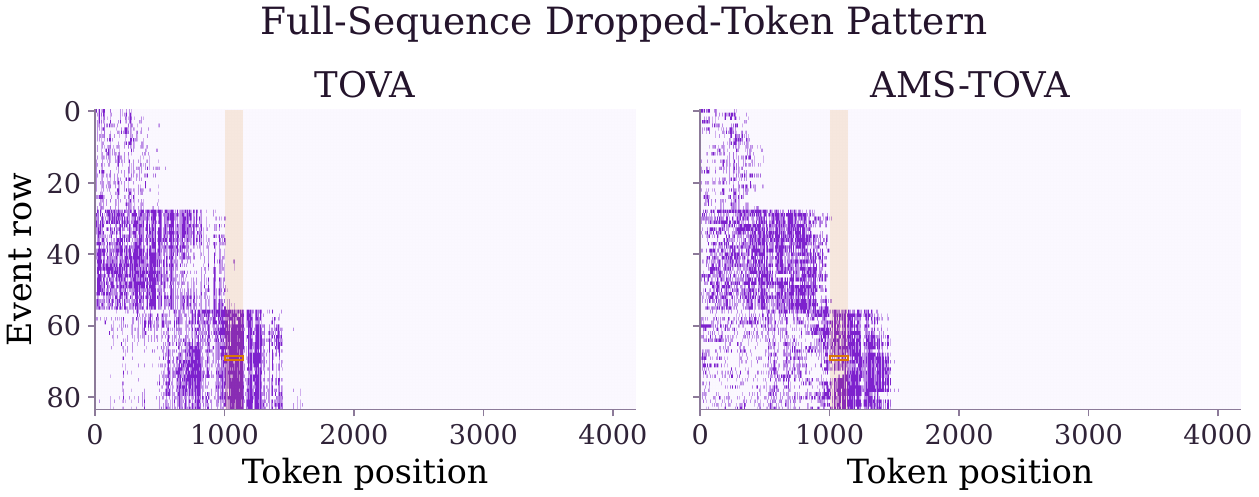}
    \caption{Full-sequence view.}
    \label{fig:drop-full}
  \end{subfigure}\hfill
  \begin{subfigure}[t]{0.49\columnwidth}
    \centering
    \includegraphics[width=\linewidth]{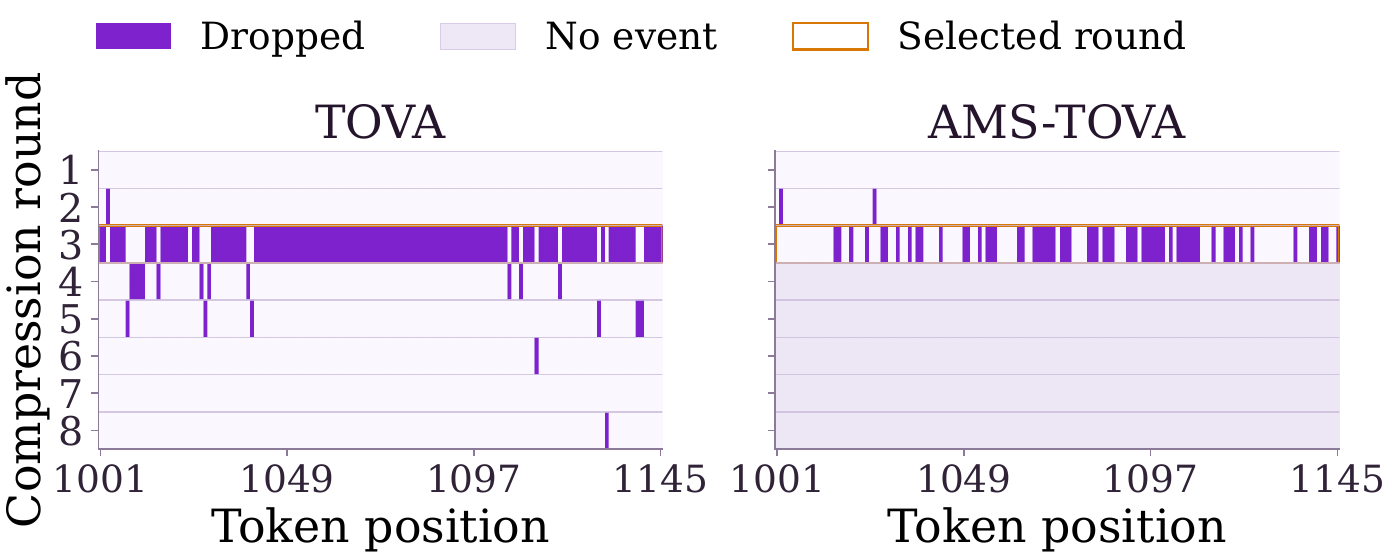} 
    \caption{Zoomed context.}
    \label{fig:drop-zoom}
  \end{subfigure}

  \caption{\textbf{Motivating dropped-token burst example.}
Purple pixels denote dropped tokens. The full-sequence view highlights a local
region where TOVA forms a dense contiguous dropped-token burst. The zoomed view
shows the same token window across compression rounds: AMS-TOVA fragments the
dropped positions in the selected round, illustrating a local failure mode that
motivates adaptive segment-wise allocation.}
  \label{fig:case-study}
\end{figure}

Figure~\ref{fig:method-overview} gives a high-level overview of the AMS pipeline. We follow the standard decoding-time KV compression paradigm: during autoregressive generation, a Transformer decoder periodically compresses its KV cache from length $T$ to a fixed target budget $T_{\mathrm{keep}}$. We compute retention policies independently for each attention head. As illustrated in Figure~\ref{fig:case-study}(a), traditional token-wise Top-$k$ policies (e.g., TOVA) tend to over-index on local attention spikes. To mitigate this, AMS replaces global Top-$k$ selection with region-wise quotas (Figure~\ref{fig:case-study}(b)), distributing retention more evenly across the temporal history without altering the backbone or the underlying scoring function. 

At each compression event, AMS keeps the backbone and scoring function unchanged, but modifies the selection stage by (i) constructing adaptive segments along the sequence within each KV head (ii) allocating token quotas to these segments under the fixed cache budget, and (iii) using a temporally smoothed mass signal to stabilize allocation. We next describe these components in more detail.

\subsection{Head-wise Quality Mass from Attention} 

At the core of our framework is a head-wise \emph{quality mass} distribution, which shifts attention signals from a token survival metric to a macro-level spatial partitioning tool. At a compression event, we collect per-position usage scores $u_i$ derived from recent-window attention over the last $W$ decoding queries (detailed implementations, including sequence pooling and causal mask padding, are deferred to Appendix~\ref{app:ams-impl}).\footnote{Attention-derived token-importance signals are widely used in training-free KV-cache eviction, including attention sinks, heavy-hitter retention, token omission, and observation-window-based KV selection~\cite{streamingllm,h2o,oren-etal-2024-transformers,snapkv}.} We then map these scores to a non-negative normalized mass distribution:

\begin{equation}
m_i \;=\; \frac{\max(u_i,0) + \epsilon}
               {\sum_{v=1}^{T} \left(\max(u_v,0) + \epsilon\right)} ,
\label{eq:mass-normalization}
\end{equation}
where $\epsilon > 0$ is a small constant. The resulting mass vector $m \in \mathbb{R}^T$ satisfies $\sum_{i=1}^T m_i = 1$. By design, $m$ assigns larger weights to regions with dense attention usage. This mass distribution subsequently drives our adaptive segmentation and quota allocation, entirely decoupled from the final token-level retention choices.

\subsection{Adaptive Mass-Segmented KV Compression}
\begin{figure}[!t]
  \centering
  \begin{subfigure}[t]{0.49\columnwidth}
    \centering
    \includegraphics[width=\linewidth]{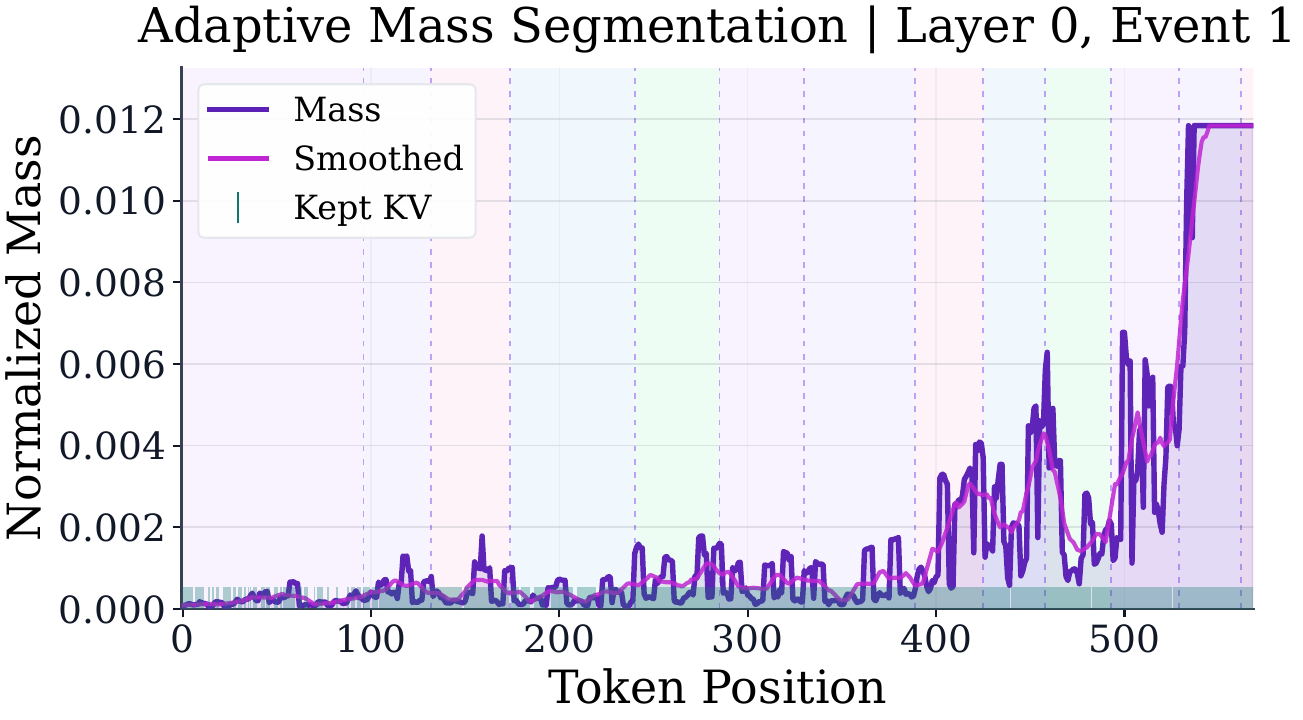}
    \caption{Compression event 1.}
    \label{fig:mass-seg-e1}
  \end{subfigure}\hfill
  \begin{subfigure}[t]{0.49\columnwidth}
    \centering
    \includegraphics[width=\linewidth]{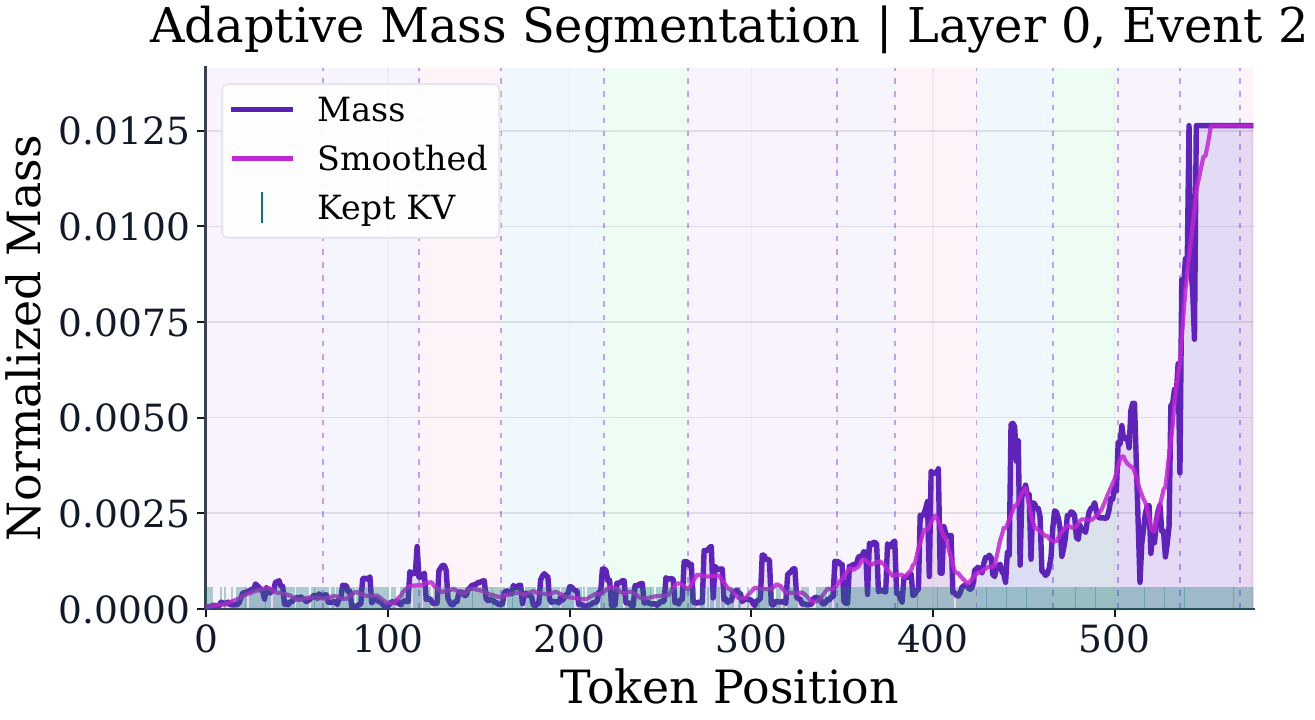}
    \caption{Compression event 2.}
    \label{fig:mass-seg-e2}
  \end{subfigure}

  \caption{\textbf{Quality mass and adaptive segmentation.}
The solid curve shows normalized quality mass over current KV-cache positions,
not absolute generation positions. Shaded bands and dashed lines denote adaptive
segments, and teal ticks mark retained KV positions. High-mass regions form
finer segments under a fixed $T_{\mathrm{keep}}$.}
  \label{fig:mass-seg}
\end{figure}

\paragraph{Mass-based adaptive segmentation.}
Given the quality-mass curve along the sequence, we partition each head's cache into adaptive segments before allocating the budget. Figure~\ref{fig:mass-seg} visualizes the normalized quality mass and the resulting adaptive segments at two consecutive compression events: regions with higher mass are cut into finer segments, while low-mass regions are merged into longer ones. By enforcing budget allocation at the segment level and guaranteeing that each segment retains at least a few tokens, we ensure that no temporal region is completely removed from the compressed cache, so every part of the history keeps at least a minimal representation.

Consider a single KV head with mass vector $m \in \mathbb{R}^T$ satisfying $\sum_{t=1}^T m_t = 1$. We first compute the prefix-sum curve $c(t) = \sum_{u=1}^{t} m_u$ for $t = 1,\dots,T$. Given a target segment mass $\Delta=\texttt{segment\_mass}$, we define cumulative-mass thresholds $\Delta, 2\Delta, 3\Delta, \dots$ and, for each threshold $k\Delta$, find the smallest position $b_k = \min \{\, t \mid c(t) \ge k\Delta \,\}$. Geometrically, this process maps the sequence into a 1D space where segment boundaries are dynamically determined by attention-mass density, fulfilling our macro-level partitioning criterion. These cut points $\{b_k\}$, together with the start and end positions, induce an initial segmentation into intervals $[a_i,b_i)$ that each contain roughly $\Delta$ total mass.

The raw cuts may produce segments that are too short or too long in terms of token count. We therefore constrain segment lengths to lie in a reasonable range by applying simple split and merge heuristics: segments longer than a length threshold $L_{\max}$ are split into roughly equal-sized subsegments, while neighboring segments shorter than $L_{\min}$ are merged. This yields a set of non-overlapping segments $\{[a_i,b_i)\}_i$ that cover all token positions $\{1,\dots,T\}$, cutting high-mass regions more finely and grouping low-mass regions more coarsely in preparation for segment-wise quota allocation.

\paragraph{Segment-wise Quota Allocation.}
Once the sequence has been partitioned into segments, we allocate the limited retention budget across segments before any token-level selection. This allocation layer enforces simple structural guarantees: every segment keeps at least a few tokens, high-mass regions receive more capacity, and globally important positions are always preserved, directly mitigating the Region Wipe-out phenomenon identified in Section~\ref{sec:intro}.

Consider a single KV head with segments $\{[a_i,b_i)\}_{i=1}^S$ and mass vector $m$. We define the mass and length of segment $i$ as $M_i = \sum_{t=a_i}^{b_i-1} m_t$ and $L_i = b_i - a_i$. Given a global keep budget $T_{\mathrm{keep}}$ and a per-segment minimum $q_{\min}$, we first assign each segment a minimum quota $q_i^{\min} = \min(q_{\min},\, L_i)$, which guarantees that no segment is completely erased as long as the global budget is large enough. The remaining budget $T_{\mathrm{rem}} = T_{\mathrm{keep}} - \sum_{i=1}^S q_i^{\min}$ is then distributed to segments in proportion to their mass as $q_i = q_i^{\min} + \mathrm{round}(T_{\mathrm{rem}} \cdot \frac{M_i}{\sum_{j=1}^S M_j})$, followed by a lightweight adjustment step to enforce $\sum_i q_i = T_{\mathrm{keep}}$ and $q_i \le L_i$. This yields a segment-wise quota vector $\{q_i\}$ that respects both the global budget and per-segment length constraints.

We additionally incorporate global must-keep rules that are standard in practice. A small number of sink tokens at the beginning of the sequence and a suffix of the most recent tokens are always retained; these positions form a must-keep set $\mathcal{I}_{\mathrm{must}}$ and take precedence when forming the final index set. If necessary, segment quotas are slightly downscaled so that the union of $\mathcal{I}_{\mathrm{must}}$ and segment-level selections fits into the overall budget. In our implementation, must-keep positions are inserted first and the remaining budget is allocated to segments; if the remaining budget becomes negative, we reduce the recent-token suffix while always preserving sink tokens.

\paragraph{In-Segment Selection with a Plug-and-Play Scorer.}
Completing our ``allocate-then-score'' paradigm, after segment-wise quotas have been determined, we perform token selection locally within each segment. This step is strictly decoupled from budget allocation: AMS only assumes access to a base scorer that assigns a real-valued importance score to each cached position, and uses the segment-wise quotas to constrain how many tokens can be kept from each region.

Formally, let $g \in \mathbb{R}^{B \times H_{kv} \times T}$ denote scores returned by a base scorer such as a TOVA-style press or an expected-attention-based scorer. For a given batch element and KV head, and for each segment $[a_i,b_i)$ with quota $q_i$, we select the top-$q_i$ indices within the segment according to $g$. This yields a set of candidate indices $\mathcal{I}_{\mathrm{seg}} = \bigcup_{i} \mathrm{TopK}\big(\{a_i,\dots,b_i-1\},\, q_i;\, g\big)$, which is then combined with the global must-keep set $\mathcal{I}_{\mathrm{must}}$ from the previous step.

We form the final keep set by taking the union $\mathcal{I} = \mathcal{I}_{\mathrm{seg}} \cup \mathcal{I}_{\mathrm{must}}$, removing duplicates, and, if necessary, trimming or backfilling a small number of positions to match the target budget $T_{\mathrm{keep}}$. Trimming is performed by discarding the lowest-scoring non-must-keep positions, while backfilling (when $\vert\mathcal{I}\vert < T_{\mathrm{keep}}$ due to short segments) adds high-scoring positions outside the current segment quotas. Finally, we apply $\mathcal{I}$ to gather the compressed cache along the sequence dimension as $K' = K[:,:, \mathcal{I},:]$ and $V' = V[:,:, \mathcal{I},:]$, and continue decoding with $(K',V')$. Because this selection step only requires a score tensor of shape $[B,H_{kv},T]$, AMS can wrap a wide range of existing scoring rules, and the mass-segmented allocation core remains compatible with diverse KV eviction methods.

\paragraph{EMA Credit for Temporal Stability.}
To stabilize quota allocations across repeated compression events, we introduce an optional exponential-moving-average (EMA) credit mechanism that makes the mass signal history-aware before segmentation and quota allocation\footnote{History-aware smoothing has been used to stabilize KV-cache eviction under repeated compression~\cite{restkv,gkv}.}.

For each layer and KV head, we maintain a credit vector $c \in \mathbb{R}^T$ that accumulates past mass assignments. At a new compression event, given the current mass $m^{(e)}$, we update $c \leftarrow \lambda \, c + (1-\lambda)\, m^{(e)}$, where $\lambda \in (0,1)$ is a decay parameter. We then normalize the credit into a distribution and form a history-aware mass $m_{\text{used}} = \mathrm{normalize}\big(\beta \, m^{(e)} + (1-\beta)\, \mathrm{normalize}(c)\big)$, with mixing coefficient $\beta \in [0,1]$. The resulting $m_{\text{used}}$ replaces $m$ in the segmentation and quota allocation steps; all subsequent logic is unchanged. Intuitively, tokens and regions that remain consistently useful across multiple events accumulate higher credit and therefore receive a larger share of the budget over time, while transient spikes in recent attention are smoothed out. In practice, this history-aware mass leads to smaller changes in segment boundaries and keep sets across compression events (Sec.~\ref{subsec:ablation}).

\section{Experiments}
\label{sec:experiments}

\subsection{Experimental Settings}
\label{subsec:exp-settings}

\paragraph{Benchmarks and models.}
We evaluate decoding-time KV compression primarily on mathematical reasoning (\textsc{Math500}~\cite{hendrycks2021math,lightman2024verify}, \textsc{AIME24}, \textsc{AIME25}~\cite{tigerlab_aime25}, and \textsc{GSM8K}~\cite{cobbe2021gsm8k}), with additional long-context workloads (e.g., code completion and open-domain QA) evaluated in Section~\ref{subsec:efficiency-generality}. Our main backbone is \textsc{DeepSeek-R1-Distill-Qwen-7B}~\cite{deepseek_r1_distill_qwen7b}, with scalability tested on its 32B variant~\cite{deepseek_r1_distill_qwen32b} and cross-backbone generalization on \textsc{OpenThinker3-7B}~\cite{openthinker3_7b}. All methods are evaluated under identical decoding-time constraints (e.g., per-layer budget $T_{\mathrm{keep}} \in \{256,512,1024\}$, interval $I=512$). We compare AMS against representative training-free KV-cache compressors, including StreamingLLM~\cite{streamingllm}, TOVA~\cite{oren-etal-2024-transformers}, KeyDiff~\cite{keydiff}, ChunkKV-Expected~\cite{chunkkv}, PyramidKV~\cite{pyramidkv}, AdaKV-ExpE2~\cite{adakv,expectedattn}, R-KV~\cite{rkv}, RPC~\cite{rpc}, and TriAttention~\cite{triattention_paper}. Main results instantiate AMS with TOVA and expected-attention scorers, yielding AMS-TOVA and AMS-Expected. Additional plug-in results, including TriAttention, KeyDiff, and R-KV, are reported in Table~\ref{tab:math500-universality}. Detailed evaluation protocols, baseline configurations, and AMS hyperparameters are deferred to Appendix~\ref{app:full-details}.

\subsection{Main Results}
\label{subsec:qwen-main}

\paragraph{Overall performance.}
We evaluate decoding-time KV compression on \textsc{Math500}, \textsc{AIME24}, and \textsc{AIME25} using \textsc{DeepSeek-R1-Distill-Qwen-7B} as the backbone.
Table~\ref{tab:math500-aime25} reports pass@1 accuracy for the main compared methods under per-layer cache budgets $T_{\mathrm{keep}} \in \{256, 512, 1024\}$. The gains across different baseline philosophies are substantial. On \textsc{Math500}, \textbf{AMS-TOVA} improves over the token-wise \textbf{TOVA} by $+7.2$, $+3.8$, and $+4.6$ absolute points at $T_{\mathrm{keep}}=256,512,1024$, respectively. The expected-attention version \textbf{AMS-Expected} is also strong, outperforming the corresponding expected-attention baseline \textbf{AdaKV-ExpE2} by $+16.0$, $+7.4$, and $+0.8$ points at these budgets. These results indicate that the proposed allocation layer improves both token-wise and expected-attention-based scoring under the same decoding-time constraints. Additional plug-in results with geometric and reasoning-specific scorers, including TriAttention, are detailed in Table~\ref{tab:math500-universality}.

\begin{table*}[t]

  \caption{Pass@1 accuracy (\%) under decoding-time KV compression with per-layer effective cache length $T_{\mathrm{keep}}\in\{256,512,1024\}$. Baselines are grouped by their underlying compression paradigm. AMS-TOVA uses TOVA as the base scorer, while AMS-Expected uses the expected-attention scorer. We discuss their absolute improvements over the corresponding baselines in the main text. Additional plug-in results with other scorers are provided in Table~\ref{tab:math500-universality}. AMS variants are placed immediately below their corresponding base scorers for paired visual comparison. Bold numbers denote the best compressed result in each column.}
  
  \label{tab:math500-aime25}
  \centering
  \small
  \renewcommand{\arraystretch}{1.15}
  \setlength{\tabcolsep}{1pt}

  \begin{tabularx}{\textwidth}{@{}>{\raggedright\arraybackslash}p{0.22\textwidth}*{9}{Y}@{}}
    \toprule
    \multirow{2}{*}{\textbf{Method}} 
    & \multicolumn{3}{c}{\textbf{\textsc{Math500}}} 
    & \multicolumn{3}{c}{\textbf{\textsc{AIME24}}} 
    & \multicolumn{3}{c}{\textbf{\textsc{AIME25}}} \\
    \cmidrule(lr){2-4} \cmidrule(lr){5-7} \cmidrule(l){8-10}
    & 256 & 512 & 1024 & 256 & 512 & 1024 & 256 & 512 & 1024 \\
    \midrule
    
    Full KV 
    & \multicolumn{3}{c}{52.80} 
    & \multicolumn{3}{c}{33.33} 
    & \multicolumn{3}{c}{30.00} \\
    \midrule
    
    \rowcolor{sepGray}
    \multicolumn{10}{@{}l}{\textit{\textbf{Token-level Eviction}}} \\
    
    StreamingLLM  
    & 28.00 & 41.20 & 49.60 
    & \phantom{0}0.00 & \phantom{0}0.00 & \phantom{0}3.33 
    & \phantom{0}0.00 & \phantom{0}3.33 & 13.33 \\
    
    \rowcolor{amsLavender!28}
    TOVA          
    & 29.20 & 44.60 & 48.80 
    & \phantom{0}0.00 & \phantom{0}6.67 & 16.67 
    & \phantom{0}0.00 & \phantom{0}3.33 & 13.33 \\

    \rowcolor{amsLavender!82}
    \quad\textcolor{amsPurple}{$\hookrightarrow$} AMS-TOVA      
    & 36.40
    & 48.40
    & 53.40
    & \phantom{0}0.00
    & \phantom{0}6.67
    & \best{20.00}
    & \best{\phantom{0}3.33}
    & 13.33
    & 20.00 \\
    
    KeyDiff       
    & 22.80 & 35.60 & 49.40 
    & \phantom{0}0.00 & \phantom{0}0.00 & \phantom{0}0.00 
    & \phantom{0}0.00 & \phantom{0}3.33 & 13.33 \\
    
    \rowcolor{sepGray}
    \multicolumn{10}{@{}l}{\textit{\textbf{Structure-aware Allocation (Layer / Head / Chunk)}}} \\
    
    ChunkKV-Expected 
    & 37.20 & 48.20 & 53.00 
    & \phantom{0}0.00 & \phantom{0}3.33 & 10.00 
    & \phantom{0}0.00 & \phantom{0}6.67 & 16.67 \\
    
    PyramidKV     
    & 31.60 & 42.20 & 51.80 
    & \phantom{0}0.00 & \phantom{0}0.00 & \best{20.00} 
    & \best{\phantom{0}3.33} & \phantom{0}3.33 & \phantom{0}6.67 \\
    
    \rowcolor{amsLavender!28}
    AdaKV-ExpE2   
    & 32.60 & 46.60 & 53.40 
    & \phantom{0}0.00 & \phantom{0}6.67 & 10.00 
    & \phantom{0}0.00 & \phantom{0}6.67 & 26.67 \\

    \rowcolor{amsLavender!82}
    \quad\textcolor{amsPurple}{$\hookrightarrow$} AMS-Expected  
    & \best{48.60}
    & \best{54.00}
    & \best{54.20}
    & \best{\phantom{0}6.67}
    & \best{20.00}
    & 16.67
    & \phantom{0}0.00
    & \best{23.33}
    & \best{33.33} \\
    
    \rowcolor{sepGray}
    \multicolumn{10}{@{}l}{\textit{\textbf{Reasoning-specific Compression}}} \\
    
    R-KV          
    & 37.20 & 46.40 & 52.40 
    & \phantom{0}0.00 & \phantom{0}6.67 & 10.00 
    & \best{\phantom{0}3.33} & 10.00 & 23.33 \\
    
    RPC           
    & 34.60 & 47.20 & 52.60 
    & \phantom{0}0.00 & \phantom{0}6.67 & 10.00 
    & \best{\phantom{0}3.33} & 10.00 & 23.33 \\
    
    \bottomrule
  \end{tabularx}

\end{table*}

\paragraph{Universality across diverse base scorers.} To rigorously validate AMS as a universal plug-and-play layer, we further integrate it with geometric, gradient-free, and reasoning-specific scoring paradigms. While Table~\ref{tab:math500-aime25} focuses on overall  comparisons, Table~\ref{tab:math500-universality} isolates the absolute gains brought specifically by our allocation layer on \textsc{Math500}. As shown, AMS yields consistent improvements across these diverse scorers. Most notably, under the extreme budget of $T_{\mathrm{keep}}=256$, AMS boosts KeyDiff from $22.80\%$ to $42.80\%$ (+20.00 points) and R-KV from $37.20\%$ to $42.60\%$, demonstrating the broad efficacy of decoupling region allocation from token scoring. As shown in Table~\ref{tab:scalability_32b}, AMS's structural protection scales efficiently, yielding significant gains on \textsc{DeepSeek-R1-Distill-Qwen-32B}.

\paragraph{Effect of KV budget.}
As the keep budget shrinks, the gap between AMS and the baselines widens.
On \textsc{Math500}, for example, \textbf{StreamingLLM} and \textbf{TOVA} achieve $28.0\%$ and $29.2\%$ pass@1 at $T_{\mathrm{keep}}=256$, whereas \textbf{AMS-Expected} attains $48.6\%$ at the same budget. This pattern suggests that explicit region-wise budget allocation is particularly beneficial under aggressive decoding-time compression, allowing AMS to recover much of the accuracy lost by strong token-wise baselines when the KV cache is severely constrained. We attribute this widening gap under tight budgets to the region-level minimum guarantees in AMS, which prevent any temporal span from being fully erased in a single recompression event and thus preserve intermediate reasoning states needed by later steps.

\paragraph{Results on \textsc{GSM8K}.}
We also evaluate AMS on \textsc{GSM8K}, which consists of
shorter word problems with natural-language solutions, using the same
\textsc{DeepSeek-R1-Distill-Qwen-7B} backbone and the same decoding-time
KV-budget setting. Table~\ref{tab:qwen7b-gsm8k} reports pass@1 accuracy at
target cache sizes $\{64,128\}$. At both budgets, \textbf{AMS-Expected} achieves
the best performance among the training-free baselines: it reaches $56.7\%$ at
target cache size $64$ and $65.3\%$ at $128$, outperforming the token-wise
\textbf{TOVA} baseline by $+7.8$ and $+8.2$ points, respectively. Compared to
the head-adaptive \textbf{AdaKV-ExpE2}, AMS-Expected still yields consistent
gains ($+3.2$ at $64$ and $+2.4$ at $128$), while retaining the practical
advantage of gather-based KV compression under a fixed physical cache budget.
Overall, these results indicate that region-wise allocation within each head
can also improve robustness on more natural word-problem settings once the KV
budget is moderately large.

\begin{table*}[t]
  \centering
  \small
  \renewcommand{\arraystretch}{1.08}
  \setlength{\tabcolsep}{4pt}

  \begin{minipage}[t]{0.48\textwidth}
    \vspace{0pt}
    \centering
    \caption{Pass@1 accuracy (\%) on \textsc{GSM8K} with \textsc{DeepSeek-R1-Distill-Qwen-7B}.}
    \label{tab:qwen7b-gsm8k}
    \resizebox{\linewidth}{!}{%
    \begin{tabular}{@{}lcccc@{}}
      \toprule
      \makecell[bl]{$T_{\mathrm{keep}}$} & \makecell{Streaming\\LLM} & TOVA & \makecell{AdaKV-\\ExpE2} & \makecell{AMS-\\Expected} \\
      \midrule
      64  & 46.7 & 48.9 & 53.5 & \textbf{56.7} \\
      128 & 61.8 & 57.1 & 62.9 & \textbf{65.3} \\
      \bottomrule
    \end{tabular}%
    }
  \end{minipage}
  \hfill
  \begin{minipage}[t]{0.48\textwidth}
    \vspace{0pt}
    \centering
    \caption{Pass@1 accuracy (\%) on \textsc{Math500} with \textsc{OpenThinker3-7B}.}
    \label{tab:openthinker-math500}
    \resizebox{\linewidth}{!}{%
    \begin{tabular}{@{}lcccc@{}}
      \toprule
      $T_{\mathrm{keep}}$ & \makecell{Streaming\\LLM} & TOVA & \makecell{AMS-\\TOVA} & \makecell{AMS-\\Expected} \\
      \midrule
      128 &  9.2  & 6.4  & 12.2 & \textbf{23.6} \\
      512 & 36.8 & 25.8 & 37.4 & \textbf{45.4} \\
      \bottomrule
    \end{tabular}%
    }
  \end{minipage}
\end{table*}

\begin{table*}[t]
  \centering
  \small
  \renewcommand{\arraystretch}{1.1} 
  \setlength{\tabcolsep}{2pt}
  
  \begin{minipage}[t]{0.45\linewidth}
    \centering
    \caption{\textbf{Scalability across Model Scales.} Pass@1 accuracy (\%) on \textsc{Math500}(\textsc{Qwen-32B}).}
    \label{tab:scalability_32b}
    \begin{tabularx}{\linewidth}{@{} l *{3}{Y} @{}}
      \toprule
      \multirow{2}{*}{\textbf{Method}} & \multicolumn{3}{c}{\textbf{\textsc{Qwen-32B} ($T_{\mathrm{keep}}$)}} \\
      \cmidrule(l){2-4}
      & 256 & 512 & 1024 \\
      \midrule
      Full KV  & \multicolumn{3}{c}{47.80} \\
      \midrule
      StreamingLLM  & 11.60 & 24.20 & 40.00 \\
      TOVA          & 11.40 & 25.80 & 41.40 \\
      AdaKV-ExpE2   & 16.20 & 28.80 & 40.40 \\
      R-KV          & 26.40 & 35.60 & 40.60 \\
      RPC           & 20.80 & 36.00 & 46.40 \\
      \midrule
      \rowcolor{amsviolet} 
      AMS-TOVA (Ours)      & 20.60 & 33.40 & 44.00 \\
      \rowcolor{amsviolet} 
      AMS-Expected (Ours)  & \textbf{33.00} & \textbf{43.00} & \textbf{47.00} \\
      \bottomrule
    \end{tabularx}
  \end{minipage}
  \hfill
  \begin{minipage}[t]{0.52\linewidth}
    \centering
    \caption{\textbf{Universal Plug-in Enhancements.} Pass@1 (\%) on \textsc{Math500} using 7B backbone. AMS decouples allocation from scoring, significantly boosting diverse base scorers.}
    \label{tab:math500-universality}
    \begin{tabularx}{\linewidth}{@{} l *{3}{Y} @{}}
      \toprule
      \multirow{2}{*}{\textbf{Method}} & \multicolumn{3}{c}{\textbf{\textsc{Math500} ($T_{\mathrm{keep}}$)}} \\
      \cmidrule(l){2-4}
      & 256 & 512 & 1024 \\
      \midrule
      TriAttention (Geo.) & 55.00 & 56.80 & 57.20 \\
      \rowcolor{amsviolet}
      \quad \textbf{+ AMS-TriAtt.} & \makecell{\textbf{56.20}\amsgain{1.20}} & \makecell{\textbf{60.60}\amsgain{3.80}} & \makecell{\textbf{63.60}\amsgain{6.40}} \\
      \midrule
      KeyDiff (Grad-free) & 22.80 & 35.60 & 49.40 \\
      \rowcolor{amsviolet}
      \quad \textbf{+ AMS-KeyDiff} & \makecell{\textbf{42.80}\amsgain{20.00}} & \makecell{\textbf{53.20}\amsgain{17.60}} & \makecell{\textbf{60.40}\amsgain{11.00}} \\
      \midrule
      R-KV (Reasoning) & 37.20 & 46.40 & 52.40 \\
      \rowcolor{amsviolet}
      \quad \textbf{+ AMS-R-KV} & \makecell{\textbf{42.60}\amsgain{5.40}} & \makecell{\textbf{53.60}\amsgain{7.20}} & \makecell{\textbf{60.80}\amsgain{8.40}} \\
      \bottomrule
    \end{tabularx}
  \end{minipage}
\end{table*}

\subsection{Cross-Backbone Results}
\label{subsec:openthinker3-main}

To test whether AMS generalizes beyond our primary Qwen-style backbone, we repeat decoding-time KV compression experiments on \textsc{OpenThinker3-7B}. We keep the evaluation protocol identical to Section~\ref{subsec:exp-settings}: the same compression schedule, must-keep rules (sink and recent tokens), and the same AMS hyperparameters tuned on the Qwen backbone are reused without any re-tuning. We report pass@1 on \textsc{Math500} under two representative cache budgets, $T_{\mathrm{keep}} \in \{128,512\}$.


Table~\ref{tab:openthinker-math500} shows that AMS improves robustness on \textsc{OpenThinker3-7B}. At $T_{\mathrm{keep}}=512$, \textbf{AMS-Expected} reaches $45.4\%$ pass@1, outperforming \textbf{StreamingLLM} ($36.8\%$) and \textbf{TOVA} ($25.8\%$), and coming within $0.4$ points of the no-compression reference ($45.8\%$). At $T_{\mathrm{keep}}=128$, token-wise baselines degrade sharply: \textbf{StreamingLLM} drops to $9.2\%$ and \textbf{TOVA} to $6.4\%$, while \textbf{AMS-TOVA} reaches $12.2\%$ and \textbf{AMS-Expected} reaches $23.6\%$. These results suggest that AMS transfers well across 7B-class reasoning backbones under identical decoding-time constraints.

\begin{table*}[t]
  \centering
  \small 
  \renewcommand{\arraystretch}{1.1} 
  \setlength{\tabcolsep}{2pt} 
  
  \begin{minipage}[t]{0.52\linewidth}
  \centering
  \caption{\textbf{Generality Beyond Mathematical Reasoning.} Performance across code completion, QA, \textsc{LongBench}, and \textsc{NIAH}. AMS explicitly preserves diverse structural context.}
  \label{tab:generalization_main}
  \resizebox{\linewidth}{!}{%
  \begin{tabular}{@{}lcccc@{}}
    \toprule
    \textbf{Method} & \textbf{Repo} & \textbf{TriviaQA} & \textbf{LongBench} & \textbf{NIAH} \\
    \midrule
    Full KV (No Comp.) & 24.04 & 69.34 & 19.30 & 0.3751 \\
    \midrule
    StreamingLLM       & 26.92 & 64.57 & 19.19 & 0.2113 \\
    TOVA               & 22.40 & 67.68 & 19.12 & 0.1894 \\
    PyramidKV          & 24.10 & 65.65 & 19.02 & 0.1777 \\
    AdaKV-ExpE2        & 19.12 & 43.54 & 17.58 & 0.0866 \\
    \midrule
    \rowcolor{amsviolet}
    AMS-TOVA (Ours)    & 23.64 & \textbf{70.88} & \textbf{19.30} & \textbf{0.2384} \\
    \rowcolor{amsviolet}
    AMS-Exp. (Ours)    & \textbf{27.44} & 69.47 & 19.25 & 0.2209 \\
    \bottomrule
  \end{tabular}%
  }
\end{minipage}
  \hfill
  \begin{minipage}[t]{0.45\linewidth}
    \centering
    \caption{\textbf{Ablation of AMS components} on \textsc{Math500}. We report pass@1 (\%) under decoding-time KV compression with a TOVA scorer. Red denotes absolute performance drops.}
    \label{tab:ablation_main}
    \begin{tabularx}{\linewidth}{@{}lYY@{}} 
      \toprule
      \multirow{2}{*}{\textbf{Variant}} & \multicolumn{2}{c}{\textbf{Target Cache ($T_{\mathrm{keep}}$)}} \\
      \cmidrule(l){2-3}
      & 512 & 1024 \\
      \midrule
      
      \textbf{AMS (full)}                & \textbf{48.4} & \textbf{53.4} \\
      \midrule
      w/o mass-weighted quotas  & 46.0\drop{2.4} & 52.8\drop{0.6} \\
      w/o EMA credit            & 48.0\drop{0.4} & 50.8\drop{2.6} \\
      fixed-length segments     & 48.0\drop{0.4} & 50.4\drop{3.0} \\
      \midrule
      global head-only top-$k$  & 42.8\drop{5.6} & 49.2\drop{4.2} \\
      \bottomrule
    \end{tabularx}
  \end{minipage}
\end{table*}

\subsection{Generality and System Efficiency}
\label{subsec:efficiency-generality}

As shown in Table~\ref{tab:generalization_main}, AMS is not limited to math derivations. It achieves competitive performance across code, QA, and crucially, maintains strong overall capabilities on \textsc{LongBench} and sparse retrieval (\textsc{NIAH}). Notably, AMS-Expected outperforms all baselines and even the uncompressed Full KV on code completion (27.44\%), while AMS-TOVA achieves the highest compressed score on \textsc{TriviaQA} (70.88\%). Furthermore, we evaluate the system-level memory and latency overhead of our proposed framework (Table~\ref{tab:efficiency_main}). Importantly, AMS is compatible with paged-KV frameworks like vLLM, supporting gather-and-compact execution without steady-state attention overhead, and the serving runtime directly handles the underlying block-table replacements (Appendix~\ref{app:vllm-compat}). As shown in Table~\ref{tab:free_gen_efficiency_main}, AMS effectively suppresses logical deadlocks and repetitive hallucinations in generated text, often resulting in faster end-to-end wall-clock completion despite its minor computational overhead. Detailed quantitative system profiling is provided in Appendix~\ref{app:complexity}. We also defer the detailed mechanistic diagnostics, such as visualizing the mitigation of Region Wipe-out, to Appendix~\ref{app:mechanistic-diagnostics}.

\begin{table*}[t]
  \centering
  \small
  \setlength{\tabcolsep}{2.5pt}

  \begin{minipage}[t]{0.52\linewidth}
    \vspace{0pt}
    \centering
    \caption{\textbf{Memory and latency efficiency.} Peak memory and average decoding time on \textsc{Math500}.}
    \label{tab:efficiency_main}
    \renewcommand{\arraystretch}{0.98}
    \begin{tabularx}{\linewidth}{@{} l *{4}{Y} @{}}
      \toprule
      \multirow{2}{*}{\textbf{Method}}
      & \multicolumn{2}{c}{\textbf{Mem. (GB) $\downarrow$}}
      & \multicolumn{2}{c}{\textbf{Time (s) $\downarrow$}} \\
      \cmidrule(lr){2-3} \cmidrule(l){4-5}
      & 512 & 1024 & 128 & 512 \\
      \midrule
      StreamingLLM  & \textbf{15.0} & \textbf{14.9} & 50.1 & 48.4 \\
      TOVA          & \textbf{15.0} & 14.9 & 67.1 & 52.3 \\
      PyramidKV     & 15.3 & 15.6 & 67.5 & 51.6 \\
      AdaKV-ExpE2   & 39.6 & 39.7 & 91.8 & 62.3 \\
      \midrule
      \rowcolor{amsviolet}
      AMS-Expected  & \textbf{15.0} & 14.9 & \textbf{41.4} & \textbf{44.0} \\
      \bottomrule
    \end{tabularx}
  \end{minipage}
  \hfill
  \begin{minipage}[t]{0.45\linewidth}
    \vspace{0pt}
    \centering
    \caption{Free-generation dynamics. AMS mitigates repetition collapse, reducing the repetition rate by 5.4\% compared to TOVA. This prevents degenerate loops, decreasing total generated tokens and speeding up end-to-end runtime.}
    \label{tab:free_gen_efficiency_main}
    \renewcommand{\arraystretch}{1.45}
    \begin{tabularx}{\linewidth}{@{} l *{3}{Y} @{}}
      \toprule
      \textbf{Method}
      & \makecell{\textbf{Tokens}\\$\downarrow$}
      & \makecell{\textbf{Rep.}\\\textbf{(\%) $\downarrow$}}
      & \makecell{\textbf{Time}\\\textbf{(s) $\downarrow$}} \\
      \midrule
      TOVA     & 2963.6 & 21.58 & 67.34 \\
      \rowcolor{amsviolet}
      AMS-TOVA & \textbf{2799.7} & \textbf{16.18} & \textbf{63.85} \\
      \bottomrule
    \end{tabularx}
  \end{minipage}
  \vspace{-0.6em}
\end{table*}

\subsection{Ablation Studies}
\label{subsec:ablation}

We ablate AMS components on \textsc{Math500} using \textbf{DeepSeek-R1-Distill-Qwen-7B}. Following the setup in Section~\ref{subsec:qwen-main}, we report pass@1 under decoding-time KV compression. As shown in Table~\ref{tab:ablation_main}, our design choices are crucial for maintaining reasoning performance.

AMS separates scoring from budget allocation. To verify that the gains come from explicit region-wise structural protection rather than a new scoring rule, we first replace region-wise quotas with a single global head-only top-$k$ selection. This substantially hurts performance: at $T_{\mathrm{keep}}=512$ accuracy drops from $48.4\%$ to $42.8\%$, and at $T_{\mathrm{keep}}=1024$ from $53.4\%$ to $49.2\%$. Beyond the overarching allocation paradigm, the specific mass-weighted segmentation and EMA credit are also vital. Removing mass-weighted quotas while keeping the rest of AMS intact reduces accuracy by about $2.4$ points at $T_{\mathrm{keep}}=512$ (from $48.4\%$ to $46.0\%$). Similarly, using fixed-length segments instead of mass-based adaptive segments yields a degradation, especially at larger budgets (from $53.4\%$ to $50.4\%$ at $T_{\mathrm{keep}}=1024$). Finally, disabling the EMA credit has little effect at $T_{\mathrm{keep}}=512$ but reduces accuracy by $2.6$ points at $T_{\mathrm{keep}}=1024$, indicating that history-aware allocation becomes more beneficial when the cache budget is larger and eviction decisions are less constrained.


We further test sensitivity to the compression interval, hidden-state (HS) buffer, and segmentation configuration. On DeepSeek-R1-Distill-Qwen-7B / \textsc{Math500}, AMS-Expected remains robust across these settings (Appendix~\ref{app:ablations}): once the interval reaches 256 tokens, pass@1 changes by only a few points, and a wide range of segmentation hyperparameters yields nearly identical performance.



\section{Conclusion}

We introduced AMS, an adaptive mass-segmented framework for decoding-time KV compression. AMS reframes retention as allocate-then-score: attention-derived usage first forms adaptive temporal segments and assigns region-wise quotas, after which existing scorers select tokens locally. Across mathematical reasoning, cross-backbone transfer, larger model scales, and general long-context tasks, AMS improves the accuracy--memory trade-off under constrained KV budgets. These results show that preserving regional cache structure is an effective complement to token-level importance scoring. We discuss limitations and potential societal impacts in Appendix~\ref{app:limitations-impact}.



\small
\bibliographystyle{unsrtnat}
\bibliography{ref}
\normalsize

\newpage
\appendix
\onecolumn
\section*{Appendix}

\section{Limitations and Impact Statement}
\label{app:limitations-impact}

\paragraph{Limitations.}
AMS is a training-free decoding-time allocation layer that uses attention-derived mass for adaptive segmentation. This design keeps the method lightweight and scorer-agnostic, but it also means that AMS relies on attention-derived usage as a proxy for long-range utility. Future work may explore task-aware or learned utility signals. Although our experiments cover multiple budgets, backbones, model scales, and task families, broader evaluation on additional architectures, multi-turn interactions, and production serving systems would further clarify deployment behavior. In addition, while AMS is compatible with gather-and-compact KV execution, its practical efficiency may still depend on the serving backend, batching strategy, and hardware implementation.

\paragraph{Impact Statement.}
AMS reduces KV-memory cost for long-context inference, which may improve efficiency and accessibility of long-form language-model applications. At the same time, lowering inference cost can also make it easier to deploy long-form generation systems at scale, including systems that may generate misleading, harmful, or otherwise low-quality content. AMS should therefore be used with the same safety, monitoring, evaluation, and deployment safeguards as the underlying language model. The method does not introduce new training data or new model capabilities by itself, but it can improve the efficiency of existing models, so downstream risks mainly depend on the application context and the base model being compressed.

\section{Full Experimental and Implementation Details}
\label{app:full-details}

This appendix provides the full experimental protocol omitted from the main
text. Appendix~\ref{app:datasets-protocol} describes datasets and evaluation;
Appendix~\ref{app:baseline-config} summarizes baseline configurations and
budget enforcement; Appendix~\ref{app:ams-impl} gives AMS implementation
details; Appendix~\ref{app:ablations} reports hyperparameters and sensitivity
studies; and Appendix~\ref{app:vllm-compat} discusses compatibility with
vLLM-style paged KV serving.

\subsection{Datasets and Evaluation Protocol}
\label{app:datasets-protocol}

\paragraph{Benchmarks.}
We evaluate decoding-time KV compression on four mathematical reasoning
benchmarks:

\begin{itemize}
  \item \textbf{\textsc{Math500}} is a 500-problem subset of the
  \textsc{MATH} olympiad-style benchmark, containing competition-level
  problems across algebra, number theory, combinatorics, and geometry.
  Each instance is a free-form question with a short, typically numeric
  or symbolic, final answer.

  \item \textbf{\textsc{AIME24}} and \textbf{\textsc{AIME25}} consist of 30 problems each in the style of
  the American Invitational Mathematics Examination (AIME). Each problem
  admits a single integer answer in the range $0$ to $999$, and we follow
  the standard evaluation that counts a prediction as correct if the
  final integer matches the ground-truth answer.

  \item \textbf{\textsc{GSM8K}} is a collection of grade-school math
  word problems with natural-language solutions. We use the standard
  test split and report pass@1 accuracy following the official
  evaluation script.
\end{itemize}

For all these benchmarks, we treat the problem statement as the model
input and evaluate accuracy on a single final answer per problem
(pass@1).

\paragraph{Data splits and subsampling.}
Unless otherwise noted, all main results are reported on the full
official test split of each benchmark: 500 problems for
\textsc{Math500}, 30 problems each for \textsc{AIME24} and \textsc{AIME25}, and the standard test set for \textsc{GSM8K}. In some ablation studies (Appendix~\ref{app:ablations}), we additionally report results on a randomly sampled $10\%$ subset of \textsc{Math500} for efficiency; sampling is performed once with a fixed random seed and kept fixed across methods.

\paragraph{Generality Benchmarks.}
To rigorously assess the structural preservation capabilities of AMS beyond mathematical reasoning, we additionally incorporate broad long-context evaluations. These include \textsc{LongBench} for general long-context comprehension, \textsc{RepoBench-P} for cross-file code completion, \textsc{TriviaQA} for open-domain question answering, and the \textsc{Needle-in-a-Haystack (NIAH)} stress test for scattered structural retrieval. For these generality benchmarks, we adopt their respective standard evaluation scripts and metrics (e.g., accuracy, F1 score, or retrieval success rate) under the identical decoding-time KV constraints used for our primary math evaluations.

\paragraph{Prompting and generation.}
For each benchmark, the input to the model consists of an optional
context prefix (empty for our math benchmarks) followed by the problem text and a short answer prefix that instructs the model to produce a final answer. We use the same prompt format and answer prefix for all methods on a given benchmark, and we keep the decoding configuration (maximum generation length and any model-specific decoding settings) fixed across baselines and AMS variants.
All experiments use a fixed random seed for reproducibility, and no
problem-specific tuning is performed.

\paragraph{Answer extraction and scoring.}
We follow the official evaluation scripts provided with each benchmark.
For \textsc{Math500}, \textsc{AIME24}, and \textsc{AIME25}, the scoring code extracts a single final answer from the model output (e.g., by parsing the last boxed or integer expression) and compares it to the ground-truth answer
after simple normalization (such as stripping whitespace and ignoring
formatting).
A problem is counted as correct if this normalized prediction matches
the reference answer, and we report the resulting pass@1 accuracy.

\paragraph{Compression schedule.}
All compression methods, including AMS and the baselines described in
Appendix~\ref{app:baseline-config}, are evaluated under the same
decoding-time KV-budget setting: during autoregressive decoding we
periodically recompress the KV cache every $I$ generated tokens (default
$I = 512$) and cap the per-layer cache length at
$T_{\mathrm{keep}} \in \{256, 512, 1024\}$.
The underlying backbone, prompts, and evaluation code are identical
across methods; only the policy used to choose which KV entries to
retain within the fixed budget is changed.

\begin{table*}[t]
  \caption{Baseline methods and their exact \texttt{press\_name} keys in the evaluator registry.
  All entries below refer to \texttt{DecodingPress(base\_press=...)} wrappers unless stated otherwise. Here, $I$ and $T_{\mathrm{keep}}$ denote the evaluation-time compression interval and target cache length, respectively; in all main experiments they are set by the unified evaluation controller rather than fixed by the registry defaults.
  }
  \label{tab:baseline-press-registry}
  \centering
  \small
  \renewcommand{\arraystretch}{1.08}
  \setlength{\tabcolsep}{3.2pt}

  \begin{tabularx}{\textwidth}{@{}p{0.18\textwidth} p{0.28\textwidth} >{\raggedright\arraybackslash\ttfamily\scriptsize}X@{}}
  
    \toprule
    Paper name & press\_name (exact) & Registry construction (summary) \\
    \midrule
    No-Compression & \nolinkurl{no_press} & \nolinkurl{None} (no KV compression) \\
    StreamingLLM & \nolinkurl{decoding_streaming_llm} & \nolinkurl{DecodingPress(base_press=StreamingLLMPress())} \\
    TOVA & \nolinkurl{decoding_TOVA} & \nolinkurl{DecodingPress(base_press=TOVAPress())} \\
    TriAttention & \nolinkurl{decoding_triattention} & \nolinkurl{DecodingPress(base_press=TriAttentionPress())} \\
    KeyDiff & \nolinkurl{decoding_keydiff} & \nolinkurl{DecodingPress(base_press=KeyDiffPress())} \\
    PyramidKV & \nolinkurl{decoding_pyramidkv} & \nolinkurl{DecodingPress(base_press=PyramidKVPress())} \\
    AdaKV-ExpE2 & \nolinkurl{decoding_adakv_expected_attention_e2} &
\nolinkurl{DecodingPress(base_press=AdaKVPress(ExpectedAttentionPress(epsilon=1e-2)))} \\
    ChunkKV-Expected & \nolinkurl{decoding_chunkkv_expected_attention} &
\nolinkurl{DecodingPress(base_press=ChunkKVPress(press=ExpectedAttentionPress(epsilon=1e-2), chunk_length=20), compression_interval=I, target_size=T_keep, hidden_states_buffer_size=256)} \\ 
    R-KV & \nolinkurl{decoding_rkv} &
      \nolinkurl{DecodingPress(base_press=RKVPress(window_size=8, kernel_size=7, mix_lambda=0.07, retain_ratio=0.1))} \\
    RPC & \nolinkurl{decoding_rpc} &
\nolinkurl{DecodingPress(base_press=ReasoningPathPress(window_size=32, kernel_size=7), compression_interval=I, target_size=T_keep, hidden_states_buffer_size=64)} \\ 

    \midrule
    AMS-TOVA (ours) & \nolinkurl{decoding_adaptivesegmenthead_TOVA} &
      \nolinkurl{DecodingPress(base_press=AdaptiveMassSegmentWrapperHeadPress(base_press=TOVAPress(...), ...))} \\
    AMS-Expected (ours) & \nolinkurl{decoding_adaptive_segment_head_expected_attention} &
      \nolinkurl{DecodingPress(base_press=AdaptiveMassSegmentWrapperHeadPress(base_press=ExpectedAttentionPress(epsilon=1e-2), ...))} \\
    AMS-TriAttention (ours) & \nolinkurl{decoding_adaptivesegmenthead_triattention} &
      \nolinkurl{DecodingPress(base_press=AdaptiveMassSegmentWrapperHeadPress(base_press=TriAttentionPress(), ...))} \\
    \bottomrule
  \end{tabularx}
\end{table*}

\subsection{Baseline Configurations}
\label{app:baseline-config}

In this section we summarize the KV-cache compression baselines used in the
main experiments (Table~\ref{tab:math500-aime25}) and describe the
common decoding-time configuration under which they are evaluated.

\paragraph{Decoding-time compression setting.}
All baselines we compare to operate as decoding-time KV cache
compression policies: the model first runs a standard prefill pass on
the full prompt, and then periodically recompresses the KV cache during
autoregressive decoding.
Unless otherwise stated, we trigger a compression event every
$I = 512$ generated tokens and cap the per-layer cache length at
$T_{\mathrm{keep}} \in \{256, 512, 1024\}$.
At each event, the compressor takes the current KV cache as input and
returns a subset of positions to retain; the remaining entries are
discarded from the cache.
Across all methods, we always preserve a small number of fixed sink
tokens at the beginning of the sequence and a suffix of the most recent
tokens to stabilize generation.

\paragraph{Unified evaluation protocol.}
All methods are evaluated under the same per-layer effective KV budget
$T_{\mathrm{keep}}\in\{256,512,1024\}$ with periodic recompression.
Gather-based methods (StreamingLLM, TOVA, TriAttention, KeyDiff, PyramidKV, and our
AMS variants) physically compress the cache to $T_{\mathrm{keep}}$,
while AdaKV-ExpE2 is mask-based and enforces the budget only via attention masks.

\begin{table}[t]
  \caption{Decoding-time KV-cache compression baselines and their qualitative
  behavior. All methods are training-free and share the same cache
  budget $T_{\mathrm{keep}}$ and compression interval $I$ in our experiments.}
  \label{tab:baseline-summary}
  \centering
  \renewcommand{\arraystretch}{1.10}
  \setlength{\tabcolsep}{3pt}
  \begin{tabularx}{\linewidth}{@{}l p{2.6cm} X@{}}
    \toprule
    Method & Type & Description \\
    \midrule
    No-Compression
      & reference
      & Standard decoding without any KV compression; serves as a no-compression reference. \\

    StreamingLLM
      & streaming-style eviction
      & Maintains a small set of globally important positions together with a
        sliding window of recent tokens, designed for stable long-context
        streaming generation. \\

    TOVA
      & attention-based pruning
      & Scores cached tokens using attention weights from the most recent query
        token and keeps the highest-scoring positions under the global budget. \\

    TriAttention
      & pre-RoPE geometric pruning
      & Estimates key importance by leveraging intrinsic pre-RoPE Q/K concentration to score keys according to their spatial distance preferences. \\

    KeyDiff
      & gradient-free importance
      & Uses differences between consecutive key vectors as an importance
        signal and retains tokens whose keys change most significantly. \\

    PyramidKV
      & layer-adaptive budgeting
      & Allocates larger effective budgets to lower layers and smaller budgets
        to higher layers under a global cache limit in its original formulation;
        in our experiments it is evaluated under the same per-layer cap
        $T_{\mathrm{keep}}$ as other methods.
         \\

    AdaKV-ExpE2
      & head-adaptive, masking-based
      & Allocates retention budgets adaptively across attention heads using an
        expected-attention scoring rule (with a small regularization
        coefficient); in our implementation, eviction is implemented via
        masking rather than physically removing KV entries. \\

        ChunkKV-Expected
      & fixed-segment, scorer-wrapped
      & Wraps an underlying scorer with chunk-wise token selection: it first
        computes global token scores, then ranks chunks by aggregated score and
        keeps tokens chunk by chunk to preserve local semantic coherence under
        compression. \\

    R-KV
      & reasoning-specific scoring
      & Uses a redundancy-aware score that combines recent attention importance
        with a cosine-similarity-based redundancy penalty over historical keys,
        while protecting the prompt prefix and most recent tokens. \\

    RPC
      & reasoning-specific scoring
      & Uses a reasoning-path style score computed from a window of recent
        queries, aggregates attention over the recent path, smooths scores with
        1D average pooling, and protects the prompt prefix and recent tokens. \\

    AMS-TOVA (ours)
      & segment-wise allocation
      & Wraps a TOVA-style scorer with our adaptive mass-segmented allocation:
        attention-derived quality mass is used to form segments and assign
        region-wise quotas, after which TOVA scores are applied within each
        segment. \\

    AMS-Expected (ours)
      & segment-wise allocation
      & Same as AMS-TOVA but using an expected-attention-based scorer instead
        of TOVA. \\
        
    AMS-TriAttention (ours)
      & segment-wise allocation
      & Wraps the TriAttention geometric scorer with adaptive mass-segmented allocation. \\
    \bottomrule
  \end{tabularx}
\end{table}

\paragraph{Notes on AdaKV-ExpE2.}
In our experiments, AdaKV-ExpE2 denotes an AdaKV-style head-adaptive budget
allocation instantiated with an expected-attention scorer (we use
$\epsilon=10^{-2}$).

As a result, all baselines and AMS variants are evaluated under identical
per-layer KV budgets and compression schedules.
The only difference between methods lies in how they use
attention statistics or structural heuristics to decide which tokens to
retain within the fixed budget.

\paragraph{Implementation details for reproducibility.}
For completeness, we briefly summarize how these baselines are exposed
in our codebase.
All decoding-time baselines are instantiated via a unified method name
and executed under a decoding-only controller.
This controller applies the same compression schedule ($I = 512$,
$T_{\mathrm{keep}} \in \{256,512,1024\}$) across methods and, for
scoring-based approaches, uses a fixed HS buffer of 256
tokens.
The no-compression setting is implemented by disabling the controller
entirely, thereby keeping the full KV cache throughout decoding.

\begin{table}[t]
  \centering
  \renewcommand{\arraystretch}{1.12}
  \setlength{\tabcolsep}{3pt}
  \begin{tabularx}{\linewidth}{@{}L{0.24\linewidth}L{0.34\linewidth}L{0.16\linewidth}Z@{}}
    \toprule
    Method (paper) & Code identifier & Stage & KV handling \\
    \midrule
    No-Compression & \nolinkurl{no_press} & prefill \& decode & full KV (No Compression) \\
    StreamingLLM & \nolinkurl{decoding_streaming_llm} & decode-only & gather-based \\
    TOVA & \nolinkurl{decoding_TOVA} & decode-only & gather-based \\
    TriAttention & \nolinkurl{decoding_triattention} & decode-only & gather-based \\
    KeyDiff & \nolinkurl{decoding_keydiff} & decode-only & gather-based \\
    PyramidKV & \nolinkurl{decoding_pyramidkv} & decode-only & gather-based \\
    AdaKV-ExpE2 & \nolinkurl{decoding_adakv_expected_attention_e2} & decode-only & \textbf{mask-based (no gather)} \\
    ChunkKV-Expected & \nolinkurl{decoding_chunkkv_expected_attention} & decode-only & gather-based \\
    R-KV & \nolinkurl{decoding_rkv} & decode-only & gather-based \\
    RPC & \nolinkurl{decoding_rpc} & decode-only & gather-based \\
    \bottomrule
  \end{tabularx}
  \caption{Main baselines and the corresponding identifiers used in our
  open-source implementation.
  ``decode-only'' indicates that the compressor is applied only during
  autoregressive generation, not during prefill.}
  \label{tab:baseline-keys}
\end{table}

Beyond the methods listed in Tables~\ref{tab:baseline-summary}
and~\ref{tab:baseline-keys}, our framework also includes additional
registry entries (e.g., Knorm, QFilter, and composite variants). We
focus on the baselines above in the main paper as representative,
strong training-free methods covering streaming-style eviction,
attention-based scoring, head-adaptive allocation, and layer-adaptive
allocation.

\subsection{Existing Assets and Licenses}
\label{app:assets-licenses}

We use publicly available datasets, backbone models, and baseline implementations
only for research evaluation. Table~\ref{tab:assets-licenses} summarizes the
main existing assets used in this paper. We cite the original sources in the
main paper or appendix and follow the corresponding licenses or terms of use
where publicly available. No new dataset or pretrained model is introduced in
this work.

\begin{table}[h]
\centering
\renewcommand{\arraystretch}{1.15}
\setlength{\tabcolsep}{3pt}
\begin{tabularx}{\linewidth}{@{}L{0.25\linewidth}L{0.18\linewidth}Z Z@{}}
\toprule
Asset & Type & Usage in this paper & License / terms \\
\midrule
\textsc{Math500} & Dataset & Mathematical reasoning evaluation & Public research benchmark; original source cited \\
\textsc{AIME24}/\textsc{AIME25} & Dataset & Competition-style mathematical reasoning evaluation & Public benchmark; original source cited \\
\textsc{GSM8K} & Dataset & Grade-school math word-problem evaluation & Public research benchmark; original source cited \\
\textsc{LongBench} & Dataset & Long-context generalization evaluation & Public research benchmark; original source cited \\
\textsc{RepoBench-P} & Dataset & Code-completion evaluation & Public research benchmark; original source cited \\
\textsc{TriviaQA} & Dataset & Open-domain QA evaluation & Public research benchmark; original source cited \\
\textsc{Needle-in-a-Haystack} & Evaluation protocol & Sparse retrieval stress test & Public evaluation protocol; original source cited \\
\textsc{DeepSeek-R1-Distill-Qwen-7B/32B} & Backbone model & Main and scalability experiments & Public model; original license or model card terms followed \\
\textsc{OpenThinker3-7B} & Backbone model & Cross-backbone generalization experiments & Public model; original license or model card terms followed \\
\textsc{KvPress} & Software framework & Decoding-time KV-compression implementation & Public software package; original license followed \\
Baseline KV compressors & Algorithms / implementations & Comparative evaluation & Original papers and available implementations cited \\
\bottomrule
\end{tabularx}
\caption{Existing assets used in this paper.}
\label{tab:assets-licenses}
\end{table}

\subsection{Implementation Details and Pseudocode for AMS}
\label{app:ams-impl}

AMS is implemented as a training-free wrapper around a base scoring rule
(e.g., TOVA or an expected-attention scorer).
At each compression event and for each layer and KV head, AMS
(1) converts recent attention usage into a normalized quality-mass
distribution along the sequence,
(2) partitions the sequence into adaptive segments according to this
mass,
(3) allocates a segment-wise quota of kept tokens under a global budget,
and
(4) applies the base scores within each segment to select tokens.
An optional exponential moving average (EMA) credit makes the mass
history-aware across compression events.

\vspace{0.3em}
\paragraph{Default configuration.}
Unless otherwise stated, we use a single AMS configuration across all
datasets and models in the main experiments:
\begin{itemize}
  \item \emph{Mass construction.}
  Quality mass is derived from recent-window attention statistics over
  the last $W$ decoding queries (using the same $W$ for all layers). 
  Specifically, to reduce local noise, we apply a small 1D average-pooling kernel to the sequence of usage scores. Furthermore, for recent suffix positions that have fewer valid attending queries due to causal masking, we pad the missing observations with the maximum score observed in the window to avoid underestimating newly generated tokens. The scores are then normalized into a mass distribution.
  The EMA credit uses a decay parameter $\lambda\in(0,1)$ and mixing
  coefficient $\beta\in[0,1]$ as described in Section~\ref{sec:method}
  of the main text.
  \item \emph{Segmentation.}
  For each head, we construct adaptive segments by thresholding the
  cumulative mass curve with a target segment mass $\Delta$, then apply
  simple split and merge heuristics to keep segment lengths within
  $[L_{\min}, L_{\max}]$.
  The same $(\Delta, L_{\min}, L_{\max})$ are used for all layers.
  \item \emph{Quota allocation.}
  Each segment receives a minimum quota $q_{\min}$ and an additional
  quota proportional to its mass under the global budget
  $T_{\mathrm{keep}}$.
  We always keep a small number of sink tokens at the beginning of the
  sequence and a suffix of the most recent tokens; these must-keep
  positions are included on top of the segment-wise selection and take
  precedence when enforcing the overall budget.
  \item \emph{In-segment selection.}
  Given segment quotas, we apply the base scorer (TOVA or expected
  attention) independently within each segment and select the top-scoring
  positions up to the quota.
  The union of in-segment selections and must-keep positions is then
  trimmed or backfilled to match $T_{\mathrm{keep}}$ before gathering
  the compressed cache.
\end{itemize}
In all main results, these hyperparameters are kept fixed across
datasets and backbones. The exact hyperparameter values used in the main
experiments are summarized in Table~\ref{tab:ams-hparams}, and
Appendix~\ref{app:ablations} reports additional sensitivity studies.

\vspace{0.3em}
\paragraph{Pseudocode.}
Algorithm~\ref{alg:amskv} summarizes one AMS compression event for a
single layer and KV head.
The actual implementation vectorizes these operations over batch and
heads, but the logic is identical.

\begin{algorithm}[t]
  \caption{Adaptive Mass-Segmented KV Compression (single layer/head)}
  \label{alg:amskv}
  \begin{algorithmic}[1]
    \REQUIRE KV cache $K,V \in \mathbb{R}^{B \times H_{kv} \times T \times D}$
    \REQUIRE global keep budget $T_{\mathrm{keep}}$
    \REQUIRE recent-window usage scores for the last $W$ decoding queries
    \REQUIRE base scorer $\mathrm{BaseScore}$ returning
             $g \in \mathbb{R}^{B \times H_{kv} \times T}$
    \REQUIRE hyperparameters:
             target segment mass $\Delta$,
             minimum quota $q_{\min}$,
             segment length bounds $L_{\min},L_{\max}$,
             must-keep sizes $n_{\mathrm{sink}}$, $n_{\mathrm{last}}$,
             EMA parameters $\lambda,\beta$
    \ENSURE compressed cache $K', V'$ with length $T_{\mathrm{keep}}$

    \STATE \textbf{Step 1: quality mass from recent attention}
    \STATE aggregate usage of each cached position from the last $W$ queries
           to obtain scores $u_1,\dots,u_T$
    \STATE optionally smooth $\{u_t\}$ along the sequence (e.g., short 1D averaging)
    \STATE normalize to a non-negative mass distribution
           $m^{(\mathrm{cur})}$ via
           $m_t^{(\mathrm{cur})} \propto \max(u_t,0) + \varepsilon$,
           $\sum_t m_t^{(\mathrm{cur})}=1$

    \STATE \textbf{Step 2: history-aware mass (EMA credit)}
    \IF{credit is used}
      \STATE update credit $c \leftarrow \lambda c + (1-\lambda)\,m^{(\mathrm{cur})}$
      \STATE $\tilde{c} \leftarrow \mathrm{normalize}(c)$
      \STATE $m \leftarrow \mathrm{normalize}\big(\beta\, m^{(\mathrm{cur})}
             + (1-\beta)\,\tilde{c}\big)$
    \ELSE
      \STATE $m \leftarrow m^{(\mathrm{cur})}$
    \ENDIF

    \STATE \textbf{Step 3: mass-based adaptive segmentation}
    \STATE compute prefix sums $c(t) = \sum_{u=1}^{t} m_u$ for $t=1,\dots,T$
    \STATE define thresholds $k\Delta$ for $k=1,2,\dots$ and find cut points
           $b_k = \min\{t : c(t) \ge k\Delta\}$
    \STATE form initial segments from $\{0,\{b_k\},T\}$ as intervals $[a_i,b_i)$
    \FOR{each segment $[a_i,b_i)$}
      \IF{$b_i - a_i > L_{\max}$}
        \STATE split into roughly equal subsegments
      \ELSIF{$b_i - a_i < L_{\min}$}
        \STATE merge with a neighboring segment
      \ENDIF
    \ENDFOR

    \STATE \textbf{Step 4: segment-wise quota allocation}
    \STATE for each segment $i$, compute length $L_i = b_i-a_i$ and
           mass $M_i = \sum_{t=a_i}^{b_i-1} m_t$
    \STATE assign minimum quota $q_i^{\min} = \min(q_{\min}, L_i)$
    \STATE $T_{\mathrm{rem}} \leftarrow
           T_{\mathrm{keep}} - \sum_i q_i^{\min}$
    \STATE distribute $T_{\mathrm{rem}}$ proportionally to $\{M_i\}$:
           $q_i \leftarrow q_i^{\min}
           + \mathrm{round}\!\big(T_{\mathrm{rem}} M_i / \sum_j M_j\big)$
    \STATE clip $q_i \le L_i$ and adjust $\{q_i\}$ so that
           $\sum_i q_i = T_{\mathrm{keep}}$

    \STATE \textbf{Step 5: must-keep set and in-segment selection}
    \STATE define must-keep positions
           $\mathcal{I}_{\mathrm{must}}$ as the first
           $n_{\mathrm{sink}}$ tokens and the last
           $n_{\mathrm{last}}$ tokens (when feasible)
    \STATE obtain token scores $g = \mathrm{BaseScore}(K,V)$
    \FOR{each batch index $b$ and KV head $h$}
      \FOR{each segment $[a_i,b_i)$ with quota $q_i$}
        \STATE select top-$q_i$ indices in
               $\{a_i,\dots,b_i-1\}$ using $g_{b,h,:}$,
               forming $\mathcal{I}^{(b,h)}_{\mathrm{seg},i}$
      \ENDFOR
      \STATE $\mathcal{I}^{(b,h)}_{\mathrm{seg}} \leftarrow
             \bigcup_i \mathcal{I}^{(b,h)}_{\mathrm{seg},i}$
      \STATE $\mathcal{I}^{(b,h)} \leftarrow
             \mathcal{I}^{(b,h)}_{\mathrm{seg}}
             \cup \mathcal{I}_{\mathrm{must}}$
      \STATE remove duplicates from $\mathcal{I}^{(b,h)}$ and sort
      \STATE trim lowest-scoring non-must-keep indices if
             $|\mathcal{I}^{(b,h)}| > T_{\mathrm{keep}}$
      \STATE backfill highest-scoring remaining indices if
             $|\mathcal{I}^{(b,h)}| < T_{\mathrm{keep}}$
    \ENDFOR

    \STATE \textbf{Step 6: gather compressed cache}
    \STATE gather $K',V'$ along the sequence axis using
           $\mathcal{I}^{(b,h)}$ for all $(b,h)$
    \STATE \textbf{return} $K', V'$
  \end{algorithmic}
\end{algorithm}

\section{Additional Mechanistic Diagnostics}
\label{app:mechanistic-diagnostics}

\begin{figure}[ht]
  \centering
  \begin{subfigure}[t]{0.48\linewidth}
    \centering
    \includegraphics[width=\linewidth]{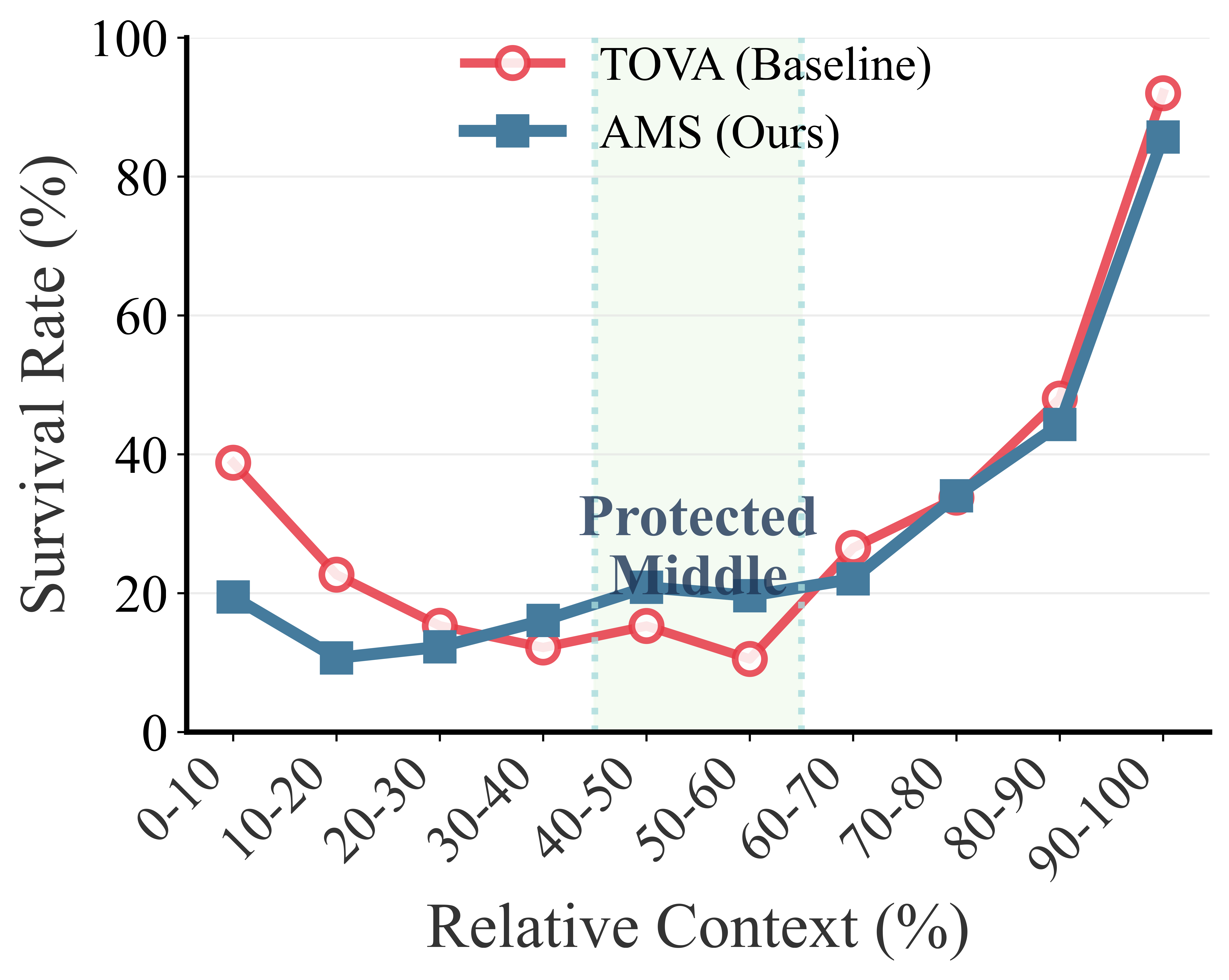} 
    \caption{Spatial retention rate}
    \label{fig:u-curve}
  \end{subfigure}
  \hfill
  \begin{subfigure}[t]{0.48\linewidth}
    \centering
    \includegraphics[width=\linewidth]{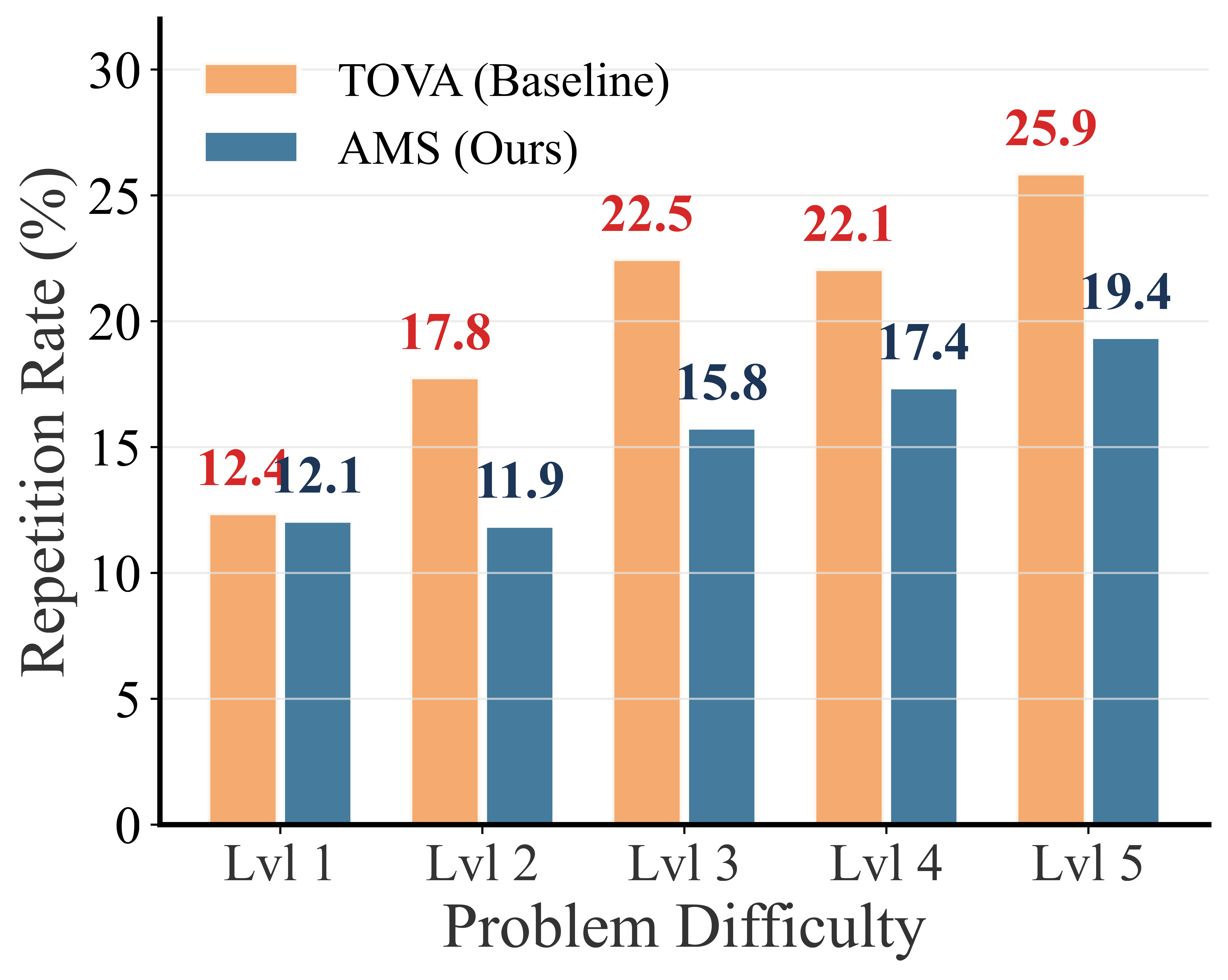}
    \caption{Repetition collapse}
    \label{fig:repetition-level}
  \end{subfigure}
  \caption{\textbf{Mechanistic insights on \textsc{Math500}.}
  (a) TOVA under-retains the middle portion of the reasoning context, while AMS
  improves middle-context coverage through segment-wise quotas.
  (b) Repetition collapse increases with problem difficulty under token-wise
  eviction; AMS suppresses this degradation.}
  \label{fig:mechanistic}
\end{figure}

\paragraph{Mitigating middle-context loss and Region Wipe-out.}
Figure~\ref{fig:mechanistic}(a) shows that token-wise selection suffers from a
strong ``lost in the middle'' pattern: TOVA under-retains tokens in the middle
portion of the reasoning context, where intermediate derivations often appear.
AMS counteracts this effect through segment-wise quotas, improving retention in
the middle region and reducing the Region Wipe-out Rate from $11.3\%$ to
$8.7\%$.

We further measure temporal stability using the retained-set IoU between
consecutive recompression events. AMS consistently improves retained-set IoU
across mathematical domains and transformer depth stages, indicating that it
preserves the reasoning skeleton more stably over time. Detailed layer-wise and
task-wise diagnostics, including the full IoU table and stability visualizations,
are provided in Table~\ref{tab:diagnostic_iou}.

\paragraph{Suppressing Repetition Collapse.}
Figure~\ref{fig:mechanistic}(b) shows that repetition collapse becomes more
severe as problem difficulty increases. Harder problems require longer
reasoning trajectories and therefore trigger more compression events, making
token-wise eviction more likely to fragment the cache. Under this stress, TOVA exhibits a substantially higher N-gram repetition rate,
reaching $25.9\%$ on Level-5 problems, while AMS reduces it to $19.4\%$.
This suggests that region-level coverage helps the model avoid local reasoning
loops and converge more reliably.

In the subset where AMS succeeds but TOVA fails, TOVA's outputs show a much
higher repetition rate and tend to continue for more tokens without reaching a
stable conclusion. In contrast, AMS produces shorter and more stable reasoning
trajectories, supporting the view that region-wise allocation improves
long-context reasoning by preventing compression-induced state fragmentation.

We provide additional diagnostics supporting the mechanistic analysis. In particular, we measure the temporal stability of retained KV states using retained-set IoU between consecutive recompression events, and report results across both mathematical subdomains and transformer depth stages.

\begin{figure}[ht]
  \centering
  \begin{subfigure}[t]{0.48\linewidth}
    \centering
    \includegraphics[width=\linewidth]{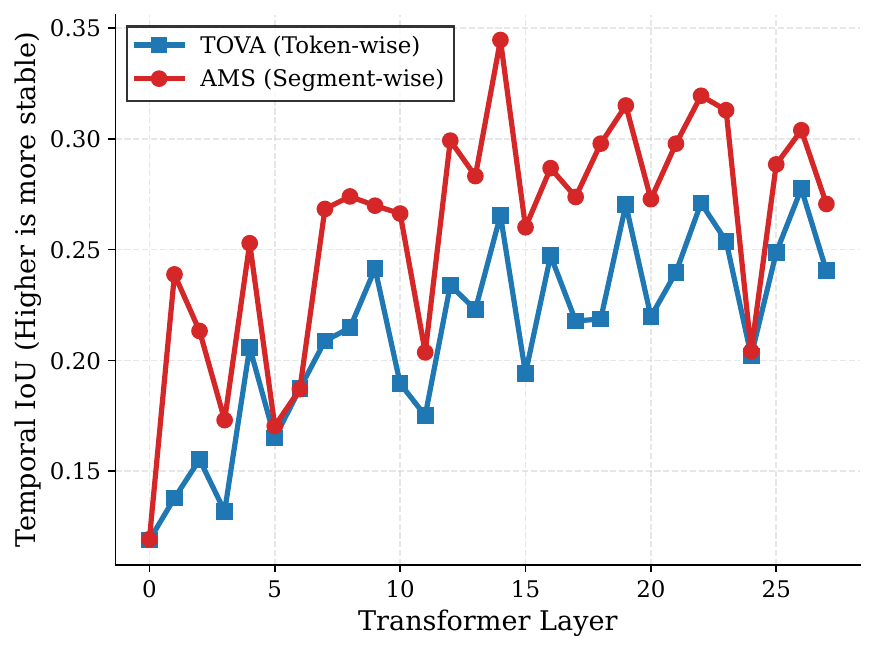} 
    \caption{Layer-wise stability}
    \label{fig:layer_iou}
  \end{subfigure}
  \hfill
  \begin{subfigure}[t]{0.48\linewidth}
    \centering
    \includegraphics[width=\linewidth]{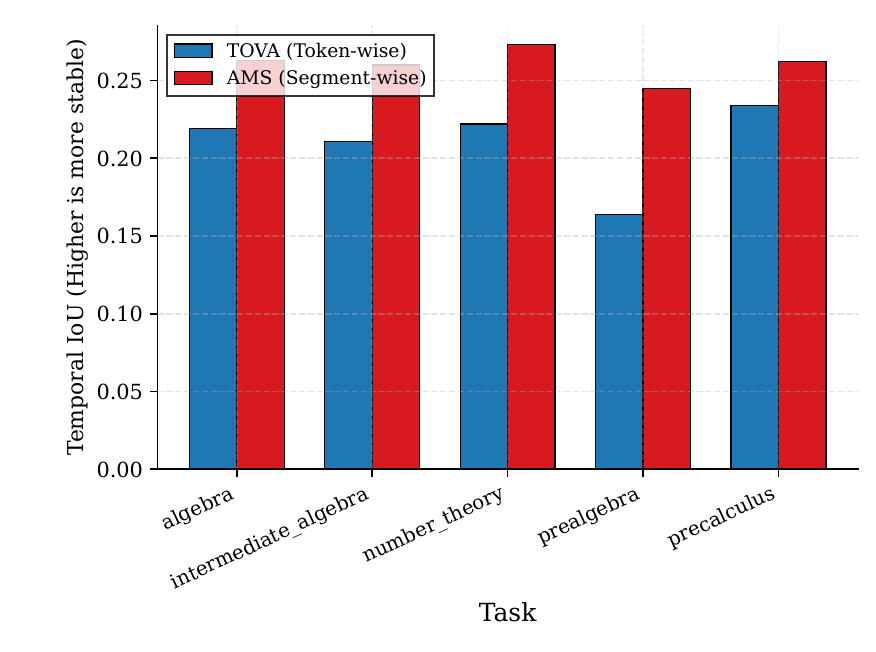} 
    \caption{Task-wise stability}
    \label{fig:task_iou}
  \end{subfigure}
  \caption{\textbf{Temporal stability of retained context.}
  AMS consistently achieves higher temporal retained-set IoU than TOVA across
  transformer layers and mathematical sub-tasks. For the token-wise TOVA
  baseline, consecutive retained tokens are grouped as proxy segments for
  direct comparison.}
  \label{fig:temporal_stability}
\end{figure}

\begin{table}[ht]
  \caption{\textbf{Additional temporal-stability diagnostics.}
  AMS improves retained-set IoU across mathematical domains and transformer
  depth stages while maintaining interpretable adaptive segments.}
  \label{tab:diagnostic_iou}
  \centering
  \renewcommand{\arraystretch}{0.95}
  \begin{tabularx}{\linewidth}{@{} l *{5}{Y} @{}}
    \toprule
    \textbf{Subset / Stage} & \textbf{Base IoU} & \textbf{AMS IoU} & \textbf{$\Delta$ IoU} & \textbf{Avg \#Segments} & \textbf{Avg Seg. Len} \\
    \midrule
    \multicolumn{6}{@{}l}{\textit{Across Mathematical Domains}} \\
    Algebra & 0.219 & 0.260 & \textbf{+0.042} & 8.05 & 138.8 \\
    Intermediate Alg. & 0.211 & 0.258 & \textbf{+0.047} & 8.80 & 138.1 \\
    Number Theory & 0.221 & 0.273 & \textbf{+0.052} & 7.61 & 135.6 \\
    Prealgebra & 0.164 & 0.244 & \textbf{+0.080} & 7.63 & 142.1 \\
    Precalculus & 0.234 & 0.262 & \textbf{+0.028} & 8.36 & 145.2 \\
    \midrule
    \multicolumn{6}{@{}l}{\textit{Across Transformer Depth}} \\
    Early (Layers 0 to 7) & 0.164 & 0.203 & \textbf{+0.039} & 10.01 & 142.4 \\
    Middle (Layers 8 to 19) & 0.224 & 0.281 & \textbf{+0.057} & 7.73 & 131.5 \\
    Late (Layers 20 to 27) & 0.244 & 0.284 & \textbf{+0.040} & 7.18 & 147.8 \\
    \bottomrule
  \end{tabularx}
\end{table}

\FloatBarrier
\section{Additional Ablations}
\label{app:ablations}

\paragraph{Overview.}
Unless otherwise stated, we use the default AMS configuration in
Table~\ref{tab:ams-hparams}. We additionally provide a sensitivity study on compression interval and HS buffer in
Table~\ref{tab:ams-interval-hsbuf}. Here \texttt{metric\_main} is the raw
metric reported by our evaluation script (see caption for details).

\begin{table}[ht]
\centering
\begin{tabularx}{\linewidth}{@{}Xc@{}}
\toprule
\textbf{Parameter} & \textbf{Value} \\
\midrule
Compression interval $I$ & 512 \\
HS buffer & 256 \\
Segment mass $\Delta$ & 0.1 \\
Min / max segment length & 16 / 256 \\
Min quota per segment $q_{\min}$ & 1 \\
EMA decay $\lambda$ & 0.9 \\
Mass mixing $\beta$ & 0.9 \\
Recent window size $W$ & 128 \\
Number of sink tokens & 4 \\
\bottomrule
\end{tabularx}

\caption{Final AMS hyperparameter configuration used for all experiments unless otherwise stated.}
\label{tab:ams-hparams}
\end{table}

\paragraph{Sensitivity to recent-window size $W$.} 
Since AMS relies on recent attention usage to compute the quality-mass distribution, we verified its sensitivity to the window size $W$. As shown in Table~\ref{tab:sensitivity_w}, on a joint sensitivity sweep over the \textsc{Math500} subset ($T_{\mathrm{keep}}=512$), AMS-Expected attained a perfectly stable pass@1 of 50.0\% for $W \in \{16, 32, 64\}$. This indicates that the EMA temporal smoothing effectively filters out instantaneous attention jitter, making the segmentation logic highly robust to the exact choice of the local attention window size.

\begin{table}[h]
  \caption{\textbf{Sensitivity to Recent-Window Size.} Ablation of the window size $W$ used for mass estimation in AMS-Expected on \textsc{Math500}. Performance remains perfectly stable.}
  \label{tab:sensitivity_w}
  \centering
  \renewcommand{\arraystretch}{0.95}
  \begin{tabularx}{0.6\linewidth}{@{} l YYY @{}}
    \toprule
    \textbf{Window Size ($W$)} & \textbf{16} & \textbf{32} & \textbf{64} \\
    \midrule
    Pass@1 (\%) & 50.0 & 50.0 & 50.0 \\
    \bottomrule
  \end{tabularx}
\end{table}

\begin{table}[t]
  \centering
  \setlength{\tabcolsep}{3pt}
  \begin{tabularx}{\linewidth}{cccccXc} 
    \toprule
    $T_{\mathrm{keep}}$ & Interval & HS buffer & \#Samples & metric\_main & sec/sample & peak GB \\
    \midrule
    \multicolumn{7}{c}{$T_{\mathrm{keep}}=512$} \\
    \midrule
    & 128  &   0 & 20 & 3112.9 & 14.3 & 14.35 \\
    & 128  &  64 & 20 & 3195.7 & 14.3 & 14.35 \\
    & 128  & 256 & 20 & 3096.6 & 14.3 & 14.35 \\
    & 256  &   0 & 24 & 2849.5 & 14.2 & 14.38 \\
    & 256  &  64 & 20 & 3070.0 & 14.2 & 14.35 \\
    & 256  & 256 & 24 & 2942.5 & 14.2 & 14.38 \\
    & 512  &   0 & 25 & 3008.7 & 14.4 & 14.40 \\
    & 512  &  64 & 24 & 3102.8 & 14.4 & 14.35 \\
    & 512  & 256 & 25 & 2856.8 & 14.3 & 14.40 \\
    & 1024 &   0 & 24 & 2857.9 & 14.3 & 14.42 \\
    & 1024 &  64 & 24 & 2758.9 & 14.1 & 14.38 \\
    & 1024 & 256 & 24 & 2977.6 & 14.3 & 14.42 \\
    \midrule
    \multicolumn{7}{c}{$T_{\mathrm{keep}}=1024$} \\
    \midrule
    & 128  &   0 & 24 & 2823.8 & 14.3 & 14.39 \\
    & 128  &  64 & 25 & 2829.7 & 14.3 & 14.38 \\
    & 128  & 256 & 24 & 2952.3 & 14.8 & 14.39 \\
    & 256  &   0 & 25 & 2770.4 & 14.3 & 14.42 \\
    & 256  &  64 & 24 & 2800.7 & 14.4 & 14.38 \\
    & 256  & 256 & 25 & 2749.0 & 14.3 & 14.42 \\
    & 512  &   0 & 26 & 2768.7 & 14.3 & 14.42 \\
    & 512  &  64 & 25 & 2647.0 & 14.1 & 14.38 \\
    & 512  & 256 & 26 & 2766.6 & 14.3 & 14.42 \\
    & 1024 &   0 & 26 & 2708.4 & 14.3 & 14.45 \\
    & 1024 &  64 & 26 & 2744.7 & 14.3 & 14.41 \\
    & 1024 & 256 & 26 & 2665.3 & 14.3 & 14.45 \\
    \midrule
    \multicolumn{7}{c}{$T_{\mathrm{keep}}=2048$} \\
    \midrule
    & 128  &   0 & 25 & 2718.1 & 14.4 & 14.43 \\
    & 128  &  64 & 25 & 2730.7 & 14.4 & 14.42 \\
    & 128  & 256 & 25 & 2754.4 & 14.6 & 14.43 \\
    & 256  &   0 & 25 & 2738.3 & 14.6 & 14.47 \\
    & 256  &  64 & 24 & 2705.3 & 14.5 & 14.43 \\
    & 256  & 256 & 25 & 2724.0 & 14.6 & 14.47 \\
    & 512  &   0 & 25 & 2770.9 & 14.6 & 14.48 \\
    & 512  &  64 & 25 & 2743.7 & 14.6 & 14.44 \\
    & 512  & 256 & 25 & 2768.8 & 14.6 & 14.48 \\
    & 1024 &   0 & 23 & 2766.5 & 14.6 & 14.51 \\
    & 1024 &  64 & 25 & 2731.5 & 14.5 & 14.47 \\
    & 1024 & 256 & 23 & 2794.9 & 14.6 & 14.51 \\
    \bottomrule
  \end{tabularx}
  \caption{Ablation study on compression interval and HS buffer for AMS-Expected on \textsc{Math500} (DeepSeek-R1-Distill-Qwen-7B, fraction=0.1). \texttt{metric\_main} indicates the raw metric output by the evaluation script.}
  \label{tab:ams-interval-hsbuf}
\end{table}

\begin{table}[t]
  \centering
  \setlength{\tabcolsep}{3.2pt}
  \renewcommand{\arraystretch}{1.05}
  \begin{tabularx}{\linewidth}{ccccccXc}
    \toprule
    $T_{\mathrm{keep}}$ & Interval & HS buffer & \#Samples & Correct & Pass@1 (\%) & sec/sample & peak GB \\
    \midrule
\multirow{12}{*}{512} & 128 & 0 & 50 & 20 & 40.0 & 62.3 & 14.35 \\
 & 128 & 64 & 50 & 20 & 40.0 & 63.9 & 14.35 \\
 & 128 & 256 & 50 & 20 & 40.0 & 61.9 & 14.35 \\
 & 256 & 0 & 50 & 24 & 48.0 & 57.0 & 14.38 \\
 & 256 & 64 & 50 & 20 & 40.0 & 61.4 & 14.35 \\
 & 256 & 256 & 50 & 24 & 48.0 & 58.9 & 14.38 \\
 & 512 & 0 & 50 & 25 & 50.0 & 60.2 & 14.40 \\
 & 512 & 64 & 50 & 24 & 48.0 & 62.1 & 14.35 \\
 & 512 & 256 & 50 & 25 & 50.0 & 57.1 & 14.40 \\
 & 1024 & 0 & 50 & 24 & 48.0 & 57.2 & 14.42 \\
 & 1024 & 64 & 50 & 24 & 48.0 & 55.2 & 14.38 \\
 & 1024 & 256 & 50 & 24 & 48.0 & 59.6 & 14.42 \\
    \midrule
\multirow{12}{*}{1024} & 128 & 0 & 50 & 24 & 48.0 & 56.5 & 14.39 \\
 & 128 & 64 & 50 & 25 & 50.0 & 56.6 & 14.38 \\
 & 128 & 256 & 50 & 24 & 48.0 & 59.0 & 14.39 \\
 & 256 & 0 & 50 & 25 & 50.0 & 55.4 & 14.42 \\
 & 256 & 64 & 50 & 24 & 48.0 & 56.7 & 14.38 \\
 & 256 & 256 & 50 & 25 & 50.0 & 55.0 & 14.42 \\
 & 512 & 0 & 50 & 26 & 52.0 & 55.4 & 14.42 \\
 & 512 & 64 & 50 & 25 & 50.0 & 52.9 & 14.38 \\
 & 512 & 256 & 50 & 26 & 52.0 & 55.3 & 14.42 \\
 & 1024 & 0 & 50 & 26 & 52.0 & 54.2 & 14.45 \\
 & 1024 & 64 & 50 & 26 & 52.0 & 54.9 & 14.41 \\
 & 1024 & 256 & 50 & 26 & 52.0 & 53.3 & 14.45 \\
    \midrule
\multirow{12}{*}{2048} & 128 & 0 & 50 & 25 & 50.0 & 54.4 & 14.43 \\
 & 128 & 64 & 50 & 25 & 50.0 & 54.6 & 14.42 \\
 & 128 & 256 & 50 & 25 & 50.0 & 55.1 & 14.43 \\
 & 256 & 0 & 50 & 25 & 50.0 & 54.8 & 14.47 \\
 & 256 & 64 & 50 & 24 & 48.0 & 56.8 & 14.43 \\
 & 256 & 256 & 50 & 25 & 50.0 & 55.9 & 14.47 \\
 & 512 & 0 & 50 & 25 & 50.0 & 55.4 & 14.48 \\
 & 512 & 64 & 50 & 25 & 50.0 & 54.9 & 14.44 \\
 & 512 & 256 & 50 & 25 & 50.0 & 55.4 & 14.48 \\
 & 1024 & 0 & 50 & 23 & 46.0 & 56.8 & 14.51 \\
 & 1024 & 64 & 50 & 25 & 50.0 & 54.6 & 14.47 \\
 & 1024 & 256 & 50 & 23 & 46.0 & 58.3 & 14.51 \\
    \bottomrule
  \end{tabularx}
  \caption{Sweep of compression interval and HS buffer for AMS-Expected on \textsc{Math500} (DeepSeek-R1-Distill-Qwen-7B, fraction=0.1). Correct = metric\_main in the csv.}
  \label{tab:ams-interval-hsbuf-full}
\end{table}

\begin{table}[t]
  \centering
  \setlength{\tabcolsep}{3.0pt}
  \renewcommand{\arraystretch}{1.05}
  \begin{tabularx}{\linewidth}{ccccccccXc}
    \toprule
    $W$ & $\Delta$ & $L_{\min}$ & $L_{\max}$ & $q_{\min}$ & keep\_last & $n_{\mathrm{sink}}$ & Pass@1 (\%) & sec/sample & peak GB \\
    \midrule
16 & 0.005 & 32 & 1024 & 32 & 16 & 4 & 50.0 & 55.9 & 14.48 \\
16 & 0.005 & 64 & 1024 & 8 & 32 & 8 & 50.0 & 54.4 & 14.48 \\
16 & 0.005 & 128 & 512 & 0 & 16 & 0 & 50.0 & 54.6 & 14.48 \\
16 & 0.010 & 16 & 4096 & 16 & 128 & 8 & 50.0 & 58.6 & 14.48 \\
16 & 0.010 & 32 & 1024 & 32 & 128 & 4 & 50.0 & 55.9 & 14.48 \\
16 & 0.010 & 64 & 2048 & 16 & 16 & 0 & 50.0 & 56.4 & 14.48 \\
16 & 0.010 & 128 & 1024 & 1 & 32 & 4 & 50.0 & 55.7 & 14.48 \\
16 & 0.020 & 16 & 512 & 32 & 16 & 4 & 50.0 & 54.9 & 14.48 \\
16 & 0.020 & 32 & 1024 & 8 & 64 & 8 & 50.0 & 55.1 & 14.48 \\
16 & 0.020 & 64 & 2048 & 16 & 32 & 0 & 50.0 & 55.0 & 14.48 \\
32 & 0.005 & 16 & 1024 & 32 & 16 & 4 & 50.0 & 56.2 & 14.48 \\
32 & 0.005 & 32 & 4096 & 8 & 128 & 8 & 50.0 & 54.0 & 14.48 \\
32 & 0.005 & 64 & 1024 & 0 & 16 & 0 & 50.0 & 54.8 & 14.48 \\
32 & 0.010 & 16 & 2048 & 16 & 32 & 8 & 50.0 & 56.0 & 14.48 \\
32 & 0.010 & 32 & 1024 & 32 & 64 & 4 & 50.0 & 54.6 & 14.48 \\
32 & 0.010 & 64 & 4096 & 16 & 16 & 0 & 50.0 & 54.4 & 14.48 \\
32 & 0.010 & 128 & 512 & 1 & 128 & 4 & 50.0 & 55.9 & 14.48 \\
32 & 0.020 & 16 & 1024 & 32 & 16 & 4 & 50.0 & 56.5 & 14.48 \\
32 & 0.020 & 32 & 512 & 8 & 64 & 8 & 50.0 & 54.9 & 14.48 \\
32 & 0.020 & 64 & 2048 & 16 & 32 & 0 & 50.0 & 55.5 & 14.48 \\
64 & 0.005 & 16 & 1024 & 32 & 16 & 4 & 50.0 & 56.5 & 14.48 \\
64 & 0.005 & 32 & 4096 & 16 & 64 & 0 & 50.0 & 54.2 & 14.48 \\
64 & 0.010 & 16 & 2048 & 8 & 32 & 8 & 50.0 & 55.7 & 14.48 \\
64 & 0.010 & 32 & 1024 & 16 & 128 & 4 & 50.0 & 54.4 & 14.48 \\
64 & 0.020 & 16 & 512 & 32 & 16 & 0 & 50.0 & 55.9 & 14.48 \\
    \bottomrule
  \end{tabularx}
  \caption{Sensitivity sweep for segmentation-related hyperparameters in AMS-Expected on \textsc{Math500} (fraction=0.1). Pass@1 is computed from metric\_main / num\_samples in the csv.}
  \label{tab:ams-seg-sweep}
\end{table}

\section{Compatibility with vLLM-Style Paged KV Serving}
\label{app:vllm-compat}

\paragraph{Motivation.}
Our main experiments are implemented through the HuggingFace/KVPress execution
path, where KV compression is naturally expressed as a gather operation over
dense per-layer KV tensors. In contrast, production serving systems such as
vLLM store the KV cache in paged physical blocks and use block tables to map
logical sequence positions to physical cache slots~\cite{kwon2023vllm}. This
paged layout is highly effective for batched serving and memory management, but
it raises a systems question for AMS: our retention policy may produce
head-specific keep sets $\mathcal{I}_{b,h}$, while a standard paged block table
maps each logical token position to a physical slot shared across KV heads.

\paragraph{Policy-layout separation.}
AMS naturally separates the retention policy from the physical cache layout.
The policy component decides which original token positions each KV head should
retain, producing keep indices
$\mathcal{I}\in\mathbb{N}^{B\times H_{kv}\times T_{\mathrm{keep}}}$.
The serving runtime is then responsible for deciding where the corresponding
KV vectors are stored in physical memory. This separation means that AMS does
not require changing its segmentation, quota allocation, or in-segment scoring
algorithm in order to target a paged-KV backend. Instead, the runtime only needs
to materialize the selected KV entries into a compact paged layout.

\paragraph{Head-wise compaction.}
A simple but lossy vLLM adaptation would force AMS to keep whole blocks or to
share one keep set across all KV heads. Although this would fit the existing
block-table abstraction, it would change the AMS policy and weaken the
head-wise region guarantees studied in this paper. We instead use a
compression-time head-wise compaction view.

Let $\pi_{\mathrm{old}}(b,p)$ denote the physical slot corresponding to the old
logical position $p$ of request $b$, and let $\pi_{\mathrm{new}}(b,t)$ denote
the physical slot of compact position $t$ after allocating replacement blocks.
For each layer, AMS compaction performs the conceptual copy

\begin{equation}
\begin{aligned}
s^{\mathrm{src}}_{b,h,t}
&= \pi_{\mathrm{old}}\bigl(b,\mathcal{I}_{b,h,t}\bigr), \\
s^{\mathrm{dst}}_{b,t}
&= \pi_{\mathrm{new}}(b,t).
\end{aligned}
\end{equation}
The KV entries are then materialized as
\begin{equation}
\begin{aligned}
K^{\mathrm{new}}\!\left[s^{\mathrm{dst}}_{b,t},h,:\right]
&\leftarrow
K^{\mathrm{old}}\!\left[s^{\mathrm{src}}_{b,h,t},h,:\right], \\
V^{\mathrm{new}}\!\left[s^{\mathrm{dst}}_{b,t},h,:\right]
&\leftarrow
V^{\mathrm{old}}\!\left[s^{\mathrm{src}}_{b,h,t},h,:\right].
\end{aligned}
\end{equation}

After this copy, the serving engine can continue using an ordinary compact
block table and the standard attention read path. The head-wise selection has
already been materialized into the KV contents, so the steady-state attention
kernel does not need an additional per-head indirection table. For decoder-only
attention, this is equivalent to the dense gather formulation used in our
experiments, because each KV head is attended independently and all retained
entries are past tokens. For RoPE-based models, positional information is
already encoded in the cached keys before compaction; the runtime must still
keep the next generated token at the original logical sequence position rather
than treating the compact KV length as the new autoregressive position.

\paragraph{Paged-KV runtime compaction.}
An end-to-end vLLM implementation would add a request-level compaction routine
to the KV-cache manager. At a compression event, the serving runtime would:
\begin{enumerate}[leftmargin=*]
  \item materialize the current per-request KV view needed by the AMS selector;
  \item call the AMS/KVPress selector to obtain head-wise keep indices
  $\mathcal{I}\in\mathbb{N}^{B\times H_{kv}\times T_{\mathrm{keep}}}$;
  \item allocate compact replacement blocks from the paged KV block pool;
  \item launch a layout-aware GPU copy kernel that performs the per-head KV
  movement above for every attention layer;
  \item replace the request's block-table row with the compact block IDs and
  free the old blocks after the copy completes; and
  \item maintain separate bookkeeping for the logical decoding position and the
  compact physical KV length.
\end{enumerate}
The last item is important because the compact cache length becomes
$T_{\mathrm{keep}}$, while the next generated token should still follow the
original autoregressive position.

\paragraph{Current implementation status.}
The supplementary code implements this policy--layout contract in two levels.
First, a standalone reference adapter constructs head-wise source and destination
slot mappings and verifies that attention over the compacted cache matches dense
attention over AMS-gathered KV tensors. Second, we provide a vLLM-style runtime
path that connects AMS selection to paged-KV compaction through runtime hooks:
AMS produces head-wise keep indices, the runtime compacts the physical KV blocks,
updates the request block table, and reclaims old blocks before standard vLLM
attention continues over the compact cache.

The runtime path supports decoupled mass and selection scores, including
TOVA-style and expected-attention selection. This implementation is not an
upstream vLLM patch; it is a runtime integration used to validate that AMS can
preserve head-wise selections in a paged-KV serving layout without introducing
per-head block tables in steady-state attention. We therefore report the main
benchmark results using the unified decoding-time KV-compression controller, and
use this vLLM path as an implementation and compatibility validation rather than
as a fully optimized production-serving benchmark.

\section{Hardware and Runtime Environment}
\label{app:hardware}

All experiments were conducted on Linux servers equipped with NVIDIA A800 GPUs.
Each GPU has 80 GB memory, and each evaluation job used 2 GPU(s)
unless otherwise noted.

Our implementation is based on PyTorch~2.3.1 with CUDA~12.1 and the
\textsc{KvPress} framework (v0.4.0) for decoding-time KV-cache compression.
We use the HuggingFace \textsc{Transformers} library (v4.57.3) together with
\textsc{Accelerate} (v1.0.1) for model loading and inference, and standard
scientific Python libraries including NumPy~2.3.5, SciPy~1.16.3, and
pandas~2.3.3.
Benchmark data are loaded via the \textsc{datasets} library (v2.21.0), and
plots are generated using \textsc{matplotlib} (v3.10.7).
Unless otherwise noted, all reported results are obtained under this runtime
environment using the unified decoding-time KV-compression controller
described in Appendix~\ref{app:baseline-config}. Our vLLM compatibility validation was conducted in an isolated CUDA-12.4
environment using a vLLM V1 runtime, without changing the main experimental
environment.

\clearpage

\section{Qualitative Case Studies on Failure Modes}
\label{app:case-studies}

To concretely illustrate the failure modes discussed in Section~\ref{sec:intro} and Section~\ref{app:mechanistic-diagnostics}, we provide qualitative examples from the \textsc{Math500} benchmark, comparing the generation trajectories of the baseline TOVA and our proposed AMS framework under the same KV budget ($T_{\mathrm{keep}}=512$). 

\subsection{Case 1: Problem Drifting and Repetition Collapse}
When the context budget is strictly token-wise, the model often drops the original problem constraints during long derivations. As shown in Table~\ref{tab:case_drifting}, TOVA initially attempts the polynomial division but eventually suffers from \textit{Region Wipe-out}. Consequently, it forgets the original constant ($-3$), hallucinates a new problem (changing the coefficient to $-1$), and falls into a severe \textit{repetition collapse}, indefinitely looping over the wrong equation. In contrast, AMS safely preserves the structural anchors, correctly identifying the coefficients and solving the problem steadily.

\begin{table}[h]
\caption{\textbf{Case Study 1: Problem Drifting (Polynomial Division).} The baseline forgets the initial constants and drifts into a repetitive hallucination, while AMS maintains state stability.}
\label{tab:case_drifting}
\centering
\renewcommand{\arraystretch}{1.3}
\begin{tabularx}{\linewidth}{@{}p{0.48\linewidth}|p{0.48\linewidth}@{}}
\toprule
\multicolumn{2}{l}{\textbf{Question:} Find the quotient when $x^6  - 3$ is divided by $x+1...$} \\
\midrule
\textbf{Baseline (TOVA)} & \textbf{Ours (AMS)} \\
\midrule
\dots let me compute $f(-1)f(-1)f(-1)$ where $f(x) = x^6 - 1$. \newline
\textcolor{red}{\textbf{[Problem Drifting \& Repetition Collapse]}} \newline
\textcolor{red}{Wait, no, the original polynomial is $x^6 - 1x^6 - 1x^6 - 1$?}\newline
\textcolor{red}{Wait, no, the original polynomial is $x^6 - 1x^6 - 1x^6 - 1$?}\newline
\textcolor{red}{Wait, no, the original polynomial is $x^6 - 1x^6 - 1x^6 - 1$?} \dots \newline
\textit{(Degenerates into an infinite loop and fails to converge)} 
& 
\dots The polynomial $x^6 - 3x^6 - 3x^6 - 3$ can be written as: \newline
$x^6 + 0x^5 + 0x^4 + 0x^3 + 0x^2 + 0x - 3$. \newline
\textcolor{green!60!black}{\textbf{[Stable State Tracking]}} \newline
\textcolor{green!60!black}{So the coefficients are $1, 0, 0, 0, 0, 0, -3$.} \newline
\textit{(Maintains the correct anchor constants and proceeds to the final correct quotient)} \\
\bottomrule
\end{tabularx}
\end{table}

\subsection{Case 2: Premature Overturning}
Another deceptive consequence of context fragmentation is \textit{Premature Overturning}. The model may successfully deduce the correct intermediate answer, but because the supporting derivation logic was evicted from the cache, it hallucinates a contradiction. Table~\ref{tab:case_overturning} demonstrates this behavior on an algebraic simplification task. TOVA reaches the correct intermediate state ($14$) but suddenly overturns its own conclusion, outputting $0$. AMS securely holds the logical chain and confidently boxes the correct final answer.

\begin{table}[h]
\caption{\textbf{Case Study 2: Premature Overturning (Algebraic Simplification).} The baseline reaches the correct state but overturns it due to fragmented logic, whereas AMS cleanly converges.}
\label{tab:case_overturning}
\centering
\renewcommand{\arraystretch}{1.3}
\begin{tabularx}{\linewidth}{@{}p{0.48\linewidth}|p{0.48\linewidth}@{}}
\toprule
\multicolumn{2}{p{0.96\linewidth}}{\textbf{Question:} Simplify $(u+4)(u-1) - (u-3)(u+6)$.} \\
\midrule
\textbf{Baseline (TOVA)} & \textbf{Ours (AMS)} \\
\midrule
\dots the constants calculate to $14$. \newline
\textcolor{red}{\textbf{[Premature Overturning]}} \newline
\textcolor{red}{Wait, so the correct answer is zero. But in the problem statement, the user wrote: ``Simplify $(u+4)(u-1) - (u+4)(u-1)$.''} \newline
\textcolor{red}{So, the expression is $(u+4)(u-1) - (u+4)(u-1)$. So, that's zero.} \newline
\textit{(Outputs $\boxed{0}$)}
& 
\dots Now, let's combine like terms. \newline
First, the $u^2$ terms: $u^2 - u^2$. That cancels out to $0$. \newline
Next, the $u$ terms: $3u - 3u$. That also cancels out to $0$. \newline
\textcolor{green!60!black}{\textbf{[Clean Convergence]}} \newline
\textcolor{green!60!black}{Finally, the constant terms: $-4 + 18 = 14$.} \newline
\textit{(Outputs $\boxed{14}$)} \\
\bottomrule
\end{tabularx}
\end{table}

\clearpage

\clearpage
\section{Complexity and System Efficiency}
\label{app:complexity}

Compared to standard training-free KV compressors, AMS adds only lightweight bookkeeping on top of the base scorer.
For a cache of length $T$ and $H_{kv}$ heads, computing the quality mass requires a single pass over recent-window scores and a prefix sum per head, both $O(H_{kv} T)$ operations.
Segment construction via searchsorted and the split and merge heuristics is also linear in $T$.
Quota allocation consists of a few vector operations over the number of segments, which is typically one to two orders of magnitude smaller than $T$.

In practice, the dominant cost in decoding-time compression remains the base scoring function (e.g., expected-attention statistics) and the attention computation itself.
As shown in our memory evaluation (Appendix~\ref{app:memory}), AMS matches the peak memory footprint of existing KV-cache compression baselines when run with the same target cache size. Empirically, Appendix~\ref{app:latency} shows that AMS matches or slightly improves the wall-clock decoding latency of existing gather-based policies, making it a practical drop-in replacement in long-context serving stacks.

\subsection{Memory Evaluation}
\label{app:memory}
We measure peak GPU memory usage during decoding on \textsc{Math500} with
\textsc{DeepSeek-R1-Distill-Qwen-7B}, using the same evaluation protocol as in
Section~\ref{subsec:exp-settings}. For each method, we fix the per-layer target
cache length $T_{\mathrm{keep}}$ and compression interval, and record the
maximum allocated memory (in GB) over the full run. We focus on representative, strong training-free baselines covering
streaming-style eviction (StreamingLLM), token-wise attention-based scoring
(TOVA, KeyDiff), fixed-structure allocation (ChunkKV-Expected),
layer- or head-adaptive allocation (PyramidKV, AdaKV-ExpE2),
and reasoning-oriented compression or scoring (R-KV, RPC).

For all methods that explicitly compress and gather the KV cache to the same
effective length $T_{\mathrm{keep}}$ (StreamingLLM, TOVA, PyramidKV, and
AMS-Expected), the peak GPU memory lies in a narrow band around $15$\,GB at
both $T_{\mathrm{keep}}=512$ and $T_{\mathrm{keep}}=1024$. This confirms that,
once the target physical cache size is fixed, the memory footprint is
essentially determined by $T_{\mathrm{keep}}$, and the additional bookkeeping
introduced by AMS is negligible in practice.

In contrast, AdaKV-ExpE2 follows a masking-based allocation strategy that does
not physically remove KV entries. Although it enforces an effective budget
through attention masks, it keeps (almost) the full KV cache in memory and
reaches about $39$ to $40$\,GB peak memory in both settings, more than
$2.5\times$ higher than the gather-based KV-compression methods. Compared to
AdaKV-ExpE2, AMS-Expected saves roughly $25$\,GB of peak memory while
maintaining, and often improving upon, its pass@1 accuracy, making it a much more memory-efficient choice
under the same backbone and decoding setup.

\subsection{Latency Evaluation}
\label{app:latency}
Table~\ref{tab:efficiency_main} shows the average wall-clock time per sample on
\textsc{Math500} with \textsc{DeepSeek-R1-Distill-Qwen-7B} under decoding-time
KV compression at target cache sizes $T_{\mathrm{keep}} \in \{128,512\}$.
We reuse the setup from Section~\ref{subsec:exp-settings}: for each method,
we fix the compression interval and KV budget, run a full pass over the test
set, and measure the end-to-end decoding latency per example on the same
hardware.

At $T_{\mathrm{keep}}=512$, the baseline methods
\textbf{StreamingLLM}, \textbf{TOVA}, and \textbf{PyramidKV} achieve
approximately $48.5$, $52.3$, and $51.6$ seconds per sample, respectively,
while our \textbf{AMS-Expected} runs at
about $44.0$ seconds per sample. The masking-based \textbf{AdaKV-ExpE2}
 is the slowest, at roughly $62.3$ seconds per sample.
At the tighter budget $T_{\mathrm{keep}}=128$, the pattern is similar:
AMS-Expected remains the fastest method at around $41.4$ seconds per sample,
compared to $50.1$ for StreamingLLM, $67.1$ for TOVA, $67.5$ for PyramidKV,
and $91.8$ for AdaKV-ExpE2.

Overall, AMS-Expected matches or improves the latency of existing
gather-based KV-compression policies, while AdaKV-ExpE2 incurs a
substantially higher runtime cost due to its masking-based allocation and
expected-attention scoring. Together with the memory measurements in
Appendix~\ref{app:memory}, these results show that AMS achieves a strictly
better accuracy and memory trade-off without sacrificing decoding-time
efficiency.

Crucially, standard fixed-step latency metrics do not capture the full efficiency picture of long-context reasoning. As analyzed in Section~\ref{app:mechanistic-diagnostics}, token-wise methods like TOVA are highly susceptible to \textit{Repetition Collapse} when the context fragments. To quantify the actual end-to-end efficiency, we measured the generation dynamics under free generation (Table~\ref{tab:free_gen_efficiency_main}). 

Because AMS effectively suppresses logical deadlocks and infinite loops, it significantly reduces the overall repetition rate (from 21.58\% to 16.18\%). Consequently, \textbf{AMS-TOVA} produces fewer useless hallucinated tokens and achieves a \textbf{faster end-to-end free generation time} (63.85s) compared to the underlying \textbf{TOVA} baseline (67.34s). Even under a strictly fixed 500-step setting, the lightweight linear-time overhead of AMS's segmentation logic is entirely offset by its architectural efficiency, yielding no practical slowdown.

\end{document}